\def\year{2021}\relax
\documentclass[letterpaper]{article} %
\usepackage{aaai21}  %
\usepackage{times}  %
\usepackage{helvet} %
\usepackage{courier}  %
\usepackage[hyphens]{url}  %
\usepackage{graphicx} %
\usepackage{natbib}  %
\usepackage{caption} %
\usepackage{amsmath, amssymb, amsthm}
\usepackage[american]{babel}
\usepackage{url}            %
\usepackage{booktabs}       %
\usepackage{amsfonts}       %
\usepackage{nicefrac}       %
\usepackage{microtype}      %

\usepackage{textcomp}
\usepackage{enumitem}
\usepackage{multirow}
\usepackage{multicol}
\usepackage{graphicx}
\usepackage{mathtools}
\usepackage{subcaption}

\newtheorem{theorem}{Theorem}[section]

\newtheorem{lemma}{Lemma}[section]

\newcommand{\PM}[1]{\scriptsize{\(\pm{#1}\)}}

\title{Rethinking Graph Regularization for Graph Neural Networks}
\author{
    Han Yang,
    Kaili Ma,
    James Cheng\\
}
\affiliations{
    The Chinese University of Hong Kong

    hyang@cse.cuhk.edu.hk, klma@cse.cuhk.edu.hk, jcheng@cse.cuhk.edu.hk

}

\begin{document}

\maketitle

\begin{abstract}

    The graph Laplacian regularization term is usually used in semi-supervised representation learning to provide graph structure information for a model $f(X)$. However, with the recent popularity of graph neural networks (GNNs), directly encoding graph structure $A$ into a model, i.e., $f(A, X)$, has become the more common approach. While we show that graph Laplacian regularization 
    brings little-to-no benefit to existing GNNs, and propose a simple but non-trivial variant of graph Laplacian regularization, called Propagation-regularization (P-reg), to boost the performance of existing GNN models. We provide formal analyses to show that P-reg not only infuses extra information (that is not captured by the traditional graph Laplacian regularization) into GNNs, but also has the capacity equivalent to an infinite-depth graph convolutional network. We demonstrate that P-reg can effectively boost the performance of existing GNN models on both node-level and graph-level tasks  across many different datasets.
\end{abstract}

\section{Introduction}
Semi-supervised node classification is one of the most popular and important problems in graph learning. Many effective methods~\cite{zhu2003semi,zhou2003learning,belkin2006manifold,ando2007learning}  have been proposed for node classification by adding a regularization term, e.g., Laplacian regularization, to a feature mapping model \(f(X):\mathbb{R}^{N\times F} \rightarrow \mathbb{R}^{N\times C}\), where $N$ is the number of nodes, $F$ is the dimensionality of a node feature, $C$ is the number of predicted classes, and \(X \in \mathbb{R}^{N\times F}\) is the node feature matrix. One known drawback of these methods is that the model \(f\) itself only models the features of each node in a graph and does not consider the relation among the nodes. They rely on the regularization term to capture graph structure information based on the assumption that neighboring nodes are likely to share the same class label. However, this assumption does not hold in many real world graphs as relations between nodes in these graphs could be complicated, as pointed out in~\cite{kipf2017semi}. This motivates the development of early \textit{graph neural network} (\textit{GNN}) models such as \textit{graph convolutional network} (\textit{GCN})~\cite{kipf2017semi}.

Many GNNs~\cite{kipf2017semi,velickovic2018graph,felix2019simplifying,hamilton2017inductive,pei2020geomgcn,li2015gated} have been proposed, which encode graph structure information directly into their model as \(f(A, X): (\mathbb{R}^{N \times N}, \mathbb{R}^{N\times F}) \rightarrow \mathbb{R}^{N\times C}\), where \(A \in \mathbb{R}^{N\times N}\) is the adjacency matrix of a graph. %
Then, they simply train the model by minimizing the supervised classification loss, without using graph regularization. However, in this work we ask the question: \textit{Can graph regularization also boost the performance of existing GNN models  as it does for traditional node classification models? }

We give an affirmative answer to this question. We show that existing GNNs already capture graph structure information that traditional graph Laplacian regularization can offer. Thus, we propose a new graph regularization, \textit{Propagation-regularization} (\textit{P-reg}), which is a variation of graph Laplacian-based regularization that provides new supervision signals to nodes in a graph. In addition, we prove that P-reg possesses the equivalent power as an infinite-depth GCN, which means that P-reg enables each node to capture information from nodes farther away (as a deep GCN does, but more flexible to avoid over-smoothing and much less costly to compute). We validate by experiments the effectiveness of P-reg as a general tool for boosting the performance of existing GNN models.  We believe our work could inspire a new direction for GNN framework designs.

\section{Propagation-Regularization}\label{sec:model}

\begin{figure*}[t]
  \centering
  \includegraphics[width=0.95\textwidth]{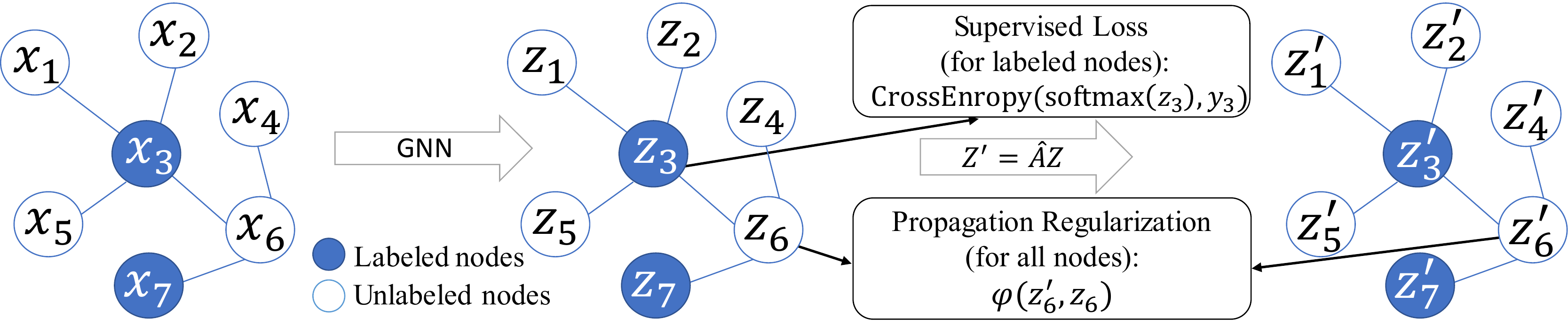}
  \caption{An overview of Propagation-regularization}
  \label{fig:overview}
\end{figure*}

The notations used in this paper and their description are also given in Appendix~A.
We use a 2-layer GCN model \(f_1\) as an example of GNN for convenience of presentation. A GCN model \(f_1\) can be formulated as 
\(
  f_1\left(A, X\right) = \hat{A}(\sigma(\hat{A}XW_0))W_1
\), where \(W_0 \in \mathbb{R}^{F \times H}\) and \(W_1 \in \mathbb{R}^{H \times C}\) are linear mapping matrices,  and \(H\) is the size of hidden units. \(\hat{A} = D^{-1}A\) is the normalized adjacency matrix, where \(D\in \mathbb{R}^{N\times N}\) is the diagonal degree matrix with \(D_{ii}=\sum_{j=1}^N A_{ij}\) and \(D_{ij} =0\) if \(i \neq j\). \ \ \(\sigma\) is the activation function.  \(f_1\) takes the graph structure and node features as input, then outputs \(Z = f_1(A, X)\in \mathbb{R}^{N \times C}\). Denote  
\(
  P_{ij} = \frac{\exp(Z_{ij})}{\sum_{k =1}^C\exp(Z_{ik})} \text{ for } i = 1,\ldots, N \text{ and } j = 1 ,\ldots, C
\) as the \textit{softmax} of the output logits.
Here, \(P \in \mathbb{R}^{N \times C}\) is the predicted class posterior probability for all nodes. By further propagating the output $Z$ of $f_1$, we obtain \(Z^\prime = \hat{A}Z \in \mathbb{R}^{N \times C}\). The corresponding softmax probability of \(Z^\prime\) is given as \(Q_{ij} = \frac{\exp(Z^\prime_{ij})}{\sum_{k =1}^C\exp(Z_{ik}^\prime)} \text{ for } i = 1,\ldots, N \text{ and } j = 1 ,\ldots, C\). %

The \textbf{Propagation-regularization} (\textbf{P-reg}) is defined as follows:
\begin{equation}\label{eq:preg}
    \mathcal{L}_{P\text{-}reg} = \frac{1}{N} \phi  \left(Z, \hat{A}Z\right),
\end{equation}
where $\hat{A}Z$ is the further propagated output of \(f_1\) and \(\phi\) is a function that measures the \textit{difference} between \(Z\) and \(\hat{A}Z\). We may adopt typical measures such as Square Error, Cross Entropy, Kullback{\textendash}Leibler Divergence, etc., for \(\phi\), and we list the corresponding function below, where \(\left(\cdot\right)_i^\top\) denotes the \(i\)-th row vector of the matrix \(\left(\cdot\right)\).

\begin{table}[h]
    \centering
    \resizebox{\linewidth}{!}{
    \begin{tabular}{cccc}
        \toprule
        \(\phi\) & Squared Error & Cross Entropy & KL Divergence\\
        \midrule
         & 
         \(\frac{1}{2}\sum\limits_{i=1}^N\left\|(\hat{A}Z)_i^\top-\left(Z\right)_i^\top\right\|_2^2\) &
         \(-\sum\limits_{i=1}^N \sum\limits_{j=1}^{C}P_{ij}\log Q_{ij}\) &
         \(\sum\limits_{i=1}^N \sum
         \limits_{j=1}^{C}P_{ij}\log \frac{P_{ij}}{Q_{ij}}\)
         \\
        \bottomrule
    \end{tabular}
    }
\end{table}

Then the GNN model \(f_1\) can be trained through the composition loss:
\begin{equation}\label{eq:preg-gnn}
  \begin{aligned}
    \mathcal{L} &= \mathcal{L}_{cls} + \mu \mathcal{L}_{P\text{-}reg} \\
    &= \text{ \small \(- \frac{1}{M} \sum_{i \in S_\text{train} } \sum_{j=1,\ldots, C} Y_{ij} \log\left(P_{ij}\right) + \mu \frac{1}{N}  \phi(Z, \hat{A}Z) \)},
  \end{aligned}
\end{equation}
where  \(S_\text{train}\) is the set of training nodes and \(M = |S_\text{train}|\), \(N\) is the number of all nodes, \(Y_{ij}\) is the ground-truth one-hot label, i.e., \(Y_{ij} = 1\) if the label of node~\(v_i\) is \(j\) and \(Y_{ij}=0\) otherwise. The \textit{regularization factor} \(\mu \geq 0\) is used to adjust the influence ratio of the two loss terms \(\mathcal{L}_{cls}\) and \(\mathcal{L}_{P\text{-}reg}\).

The computation overhead incurred by Eq.~(\ref{eq:preg}) is \(\mathcal{O}(|\mathcal{E}|C+NC)\), where \(|\mathcal{E}|\) is the number of edges in the graph. Computing \(\hat{A}Z\) has time complexity \(\mathcal{O}(|\mathcal{E}|C)\) by sparse-dense matrix multiplication or using the message passing framework~\cite{gilmer2017neural}. Evaluating \(\phi(Z, \hat{A}Z)\) has complexity \(\mathcal{O}(NC)\). 
An overview of P-reg is shown in Figure~\ref{fig:overview}. We give the annealing and thresholding version of P-reg in  Appendix~C.

\section{Understanding P-reg through Laplacian Regularization and Infinite-Depth GCN}
\label{sec:relations}

We first examine how P-reg is related to graph Laplacian regularization and infinite-depth GCN, which helps us better understand the design of P-reg and its connection to GNNs. The proofs of all lemmas and theorems are given in Appendix~B.

\subsection{Equivalence of Squared-Error P-Reg to Squared Laplacian Regularization}

Consider using squared error as $\phi$ in P-reg, i.e., \( \mathcal{L}_{P\text{-}SE} \! =\! \frac{1}{N} \phi_{SE}(Z,\! \hat{A}Z) \! =\! \frac{1}{2N}\sum_{i=1}^N\|(\hat{A}Z)_i^\top \!-\! \left(Z\right)_i^\top\|_2^2\). %
As stated by \citet{smola2003kernels}, the graph Laplacian \(\left<Z, \Delta Z\right>\) can serve as a tool to design regularization operators, where $\Delta\! =\! D\!-\!A$ is the Laplacian matrix and \(\left<\cdot, \cdot\right>\) means the inner product of two matrices. A class of regularization functions on a graph can be defined as:
\[
  \left<Z, RZ\right> \coloneqq \left<Z, r\left(\tilde{\Delta}\right)Z\right>,
\]
where \(\tilde{\Delta} = I - D^{-1}A\) is the normalized Laplacian matrix. \(r(\tilde{\Delta})\) is applying the scalar valued function \(r(\lambda)\) to eigenvalues of \(\tilde{\Delta}\) as \(r(\tilde{\Delta}) \coloneqq \sum_{i=1}^N r\left(\lambda_i\right)u_i u_i^\top \), where \(\lambda_i \text{ and } u_i\) are \(i\)-th pair of eigenvalue and eigenvector of \(\tilde{\Delta}\). 

\begin{theorem}	\label{th:sq-preg}
  The squared-error P-reg is equivalent to  the regularization with matrix \(R = \tilde{\Delta}^\top\tilde{\Delta}\).
\end{theorem}
Theorem~\ref{th:sq-preg} shows that the squared-error P-reg falls into the traditional regularization theory scope, such that we can enjoy its nice properties, e.g., constructing a Reproducing Kernel Hilbert Space \(\mathcal{H}\) by dot product \(\left<f, f\right>_{\mathcal{H}}=\left<f, \tilde{\Delta}^\top\tilde{\Delta}f\right>\), where \(\mathcal{H}\) has kernel \(\mathcal{K} = \left(\tilde{\Delta}^\top\tilde{\Delta}\right)^{-1}\). We refer readers to~\citep{smola2003kernels} for other interesting properties of spectral graph regularization and \citep{chung1997spectral} for properties of Laplacian matrix.

\subsection{Equivalence of Minimizing P-Reg to Infinite-Depth GCN}\label{sec:idgcn}

\textbf{We first analyze the behavior of infinite-depth GCN}. Take the output \(Z \in \mathbb{R}_+^{N \times C}\) of a GNN model as the input of an infinite-depth GCN. Let \(G\left(A, Z\right)\) be a graph with adjacency matrix \(A\) and node feature matrix \(Z\). A typical infinite-depth GCN applied to a graph \(G\left(A, Z\right)\) can be represented as \(\left(\hat{A}\sigma\left( \cdots \hat{A}\sigma\left(\hat{A}\sigma\left(\hat{A}ZW_0\right)W_1\right)W_2 \cdots\right)W_\infty\right) \). We use \textit{ReLU} as the \(\sigma(\cdot)\) function and neglect all linear mappings, i.e., \( W_i=I \ \  \forall i \in \mathbb{N}\). Since \(Z \in \mathbb{R}_+^{N\times C}, \text{ and ReLU}(\cdot) = \max (0, \cdot)\), the infinite-depth GCN can be simplified to \(\left(\hat{A}^\infty Z \right)\). This simplification shares similarities with SGC~\cite{felix2019simplifying}.

\begin{lemma}\label{lemma:idgcn}
	Applying graph convolution infinitely to a graph \(G\left(A, Z\right)\) with self-loops (i.e., $A:=A+I$) produces the same output vector for each node, i.e., define \(\tilde{Z}=\hat{A}^\infty Z\), then \(\tilde{z}_1=\ldots=\tilde{z}_N\), where \(\tilde{z}_i\) is the \(i\)-th row vector of \(\tilde{Z}\).
\end{lemma}
Lemma~\ref{lemma:idgcn} shows that an infinite-depth GCN makes each node capture and represent the information of the whole graph.

\noindent \textbf{We next establish the equivalence between minimizing P-reg and an infinite-depth GCN.}

\begin{lemma} \label{lemma:preg-se}
	Minimizing the squared-error P-reg of node feature matrix \(Z\) produces the same output vector for each node, i.e., \(\tilde{Z} \in \arg \min_{Z} \left\|\hat{A}Z - Z \right\|_F^2\), where \(\tilde{z}_1 \!=\!\cdots\!=\!\tilde{z}_N\).%
\end{lemma}

From Lemmas~\ref{lemma:idgcn} and~\ref{lemma:preg-se}, the behaviors of applying infinite graph convolution and minimizing the squared-error P-reg are identical. Next, we show that iteratively minimizing P-reg of \(G(A,Z)\) has the same effect as recursively applying graph convolution to \(G(A,Z)\) step by step. We can rewrite  \(\|\hat{A}Z - Z \|_F^2 = \| \left(D^{-1}A - I\right) Z\|_F^2 =  \sum_{i=1}^N \| (\frac{1}{d_i}\sum_{j \in \mathcal{N}\left(v_i\right)  }z_j) - z_i  \|_2^2 \), %
where \(d_i\) is the degree of \(v_i\) and \(\mathcal{N}\left(v_i\right)\) is the set of neighbors of \(v_i\). If we minimize \(\sum_{i=1}^N \| (\frac{1}{d_i}\sum_{j \in \mathcal{N}\left(v_i\right)  }z_j) - z_i \|_2^2 \) in an iterative manner, i.e., \(z_i^{(k+1)} = \frac{1}{d_i}\sum_{j \in \mathcal{N}\left(v_i\right)  }z_j^{(k)}\) where \(k \in \mathbb{N}\) is the iteration step, then it is exactly the same as applying a graph convolution operation to \(Z\), i.e., \(\hat{A}Z\), which can be rewritten into a node-centric representation: \(z_i^{(k+1)} = \frac{1}{d_i}\sum_{j \in \mathcal{N}\left(v_i\right)  }z_j^{(k)}\). Thus, when \(k \rightarrow +\infty\), the two formulas converge into the same point. The above analysis motivates us to establish the equivalence between minimizing different versions of P-reg and an infinite-depth GCN, giving the following theorem:

\begin{theorem}\label{theorem:preg-eq-idgcn}
  When neglecting all linear mappings in GCN, i.e., \( W_i=I \ \  \forall i \in \mathbb{N}\),  minimizing P-reg (including the three versions of $\phi$ using Squared Error, Cross Entropy and KL Divergence defined in Section~\ref{sec:model}) is equivalent to applying graph convolution infinitely to a graph \(G\left(A, Z\right)\), i.e., define \(\tilde{Z} = \hat{A}^\infty Z\), then \(\tilde{Z}\) is the optimal solution to \(\arg \min_Z \phi(Z, \hat{A}Z)\).
\end{theorem}

In Theorem~\ref{theorem:preg-eq-idgcn}, the optimal solution to minimizing P-reg corresponds to the case when \(\mu \rightarrow +\infty\) in Eq.(\ref{eq:preg-gnn}). %

\subsection{Connections to Other Existing Work}

P-reg is conceptually related to \textit{Prediction/Label Propagation}~\cite{zhu2003semi, zhou2003learning,wang2007label, karasuyama2013manifold, liu2018learning,li2019label, wang2020unifying}, \textit{Training with Soft Targets} such as distillation-based~\cite{hinton2015distilling, zhang2019your} or entropy-based~\cite{miller1996global, jaynes1957information,szegedy2016rethinking,pereyra2017regularizing,rafael2019when}, and other \textit{Graph Regularization}~\cite{civin1960linear,ding2018semi, feng2019graph, deng2019batch,verma2019graphmix, otilia2019graph} using various techniques such as Laplacian operator, adversarial training, generative adversarial networks, or co-training. 
We discuss these related works in details in Appendix~D.

\section{Why P-Reg can improve existing GNNs}
\label{sec:why}

We next answer why P-reg can boost the performance of existing GNN models. From the graph regularization perspective, P-reg provides extra information for GNNs to improve their performance. From the infinite-depth GCN perspective,  using P-reg enhances the current shallow-layered GNNs with the power of a deeper GCN as P-reg can simulate a GCN of any layers (even a \textit{fractional} layer) 
at low cost.

\subsection{Benefits of P-Reg from the Graph Regularization Perspective}	\label{why:reg}

\begin{figure*}[t]
	\centering
	\begin{subfigure}{0.33\textwidth}
		\includegraphics[width=\linewidth]{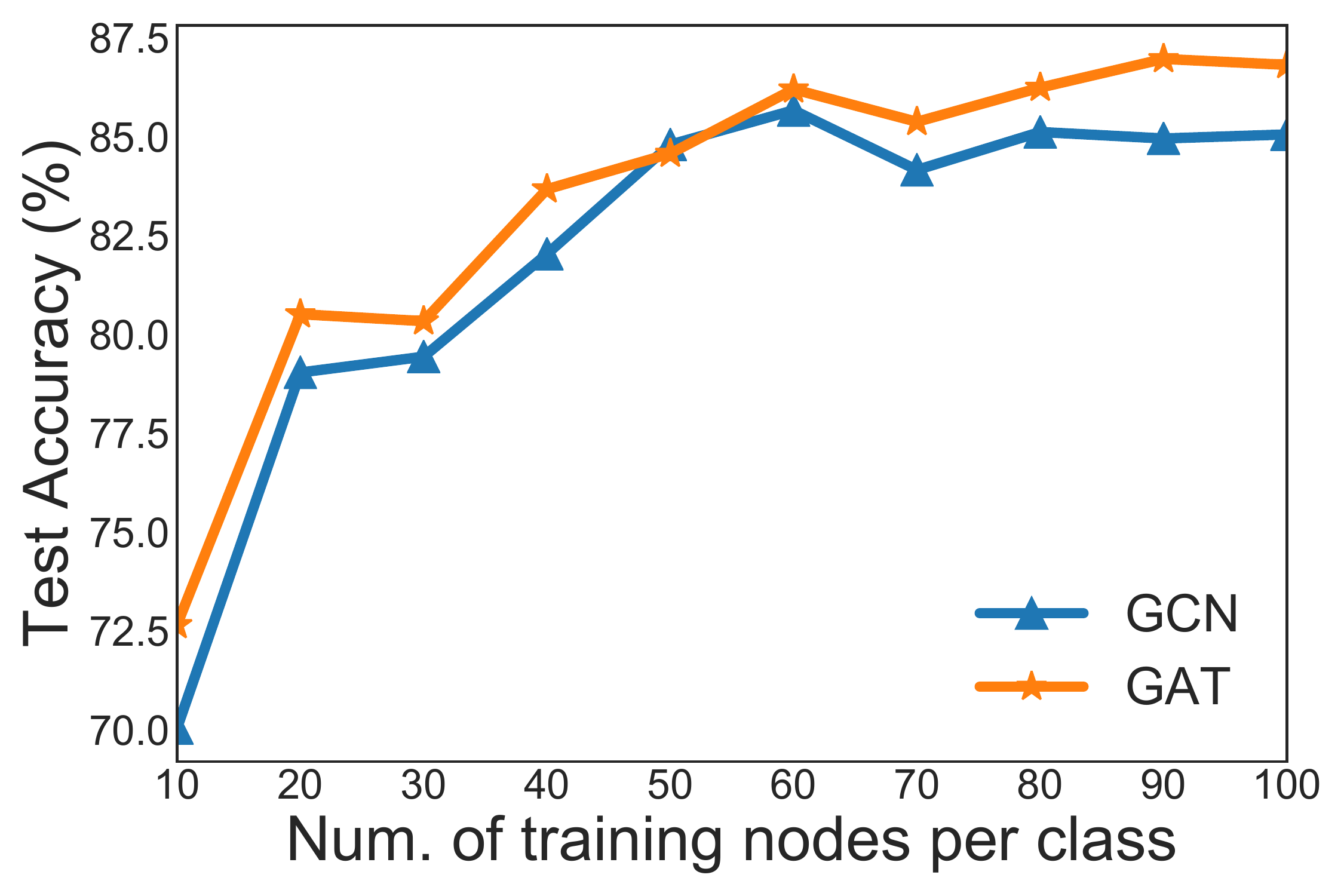}
		\caption{}
		\label{fig:training-nodes}
	\end{subfigure}
	\begin{subfigure}{0.33\textwidth}
		\includegraphics[width=\linewidth]{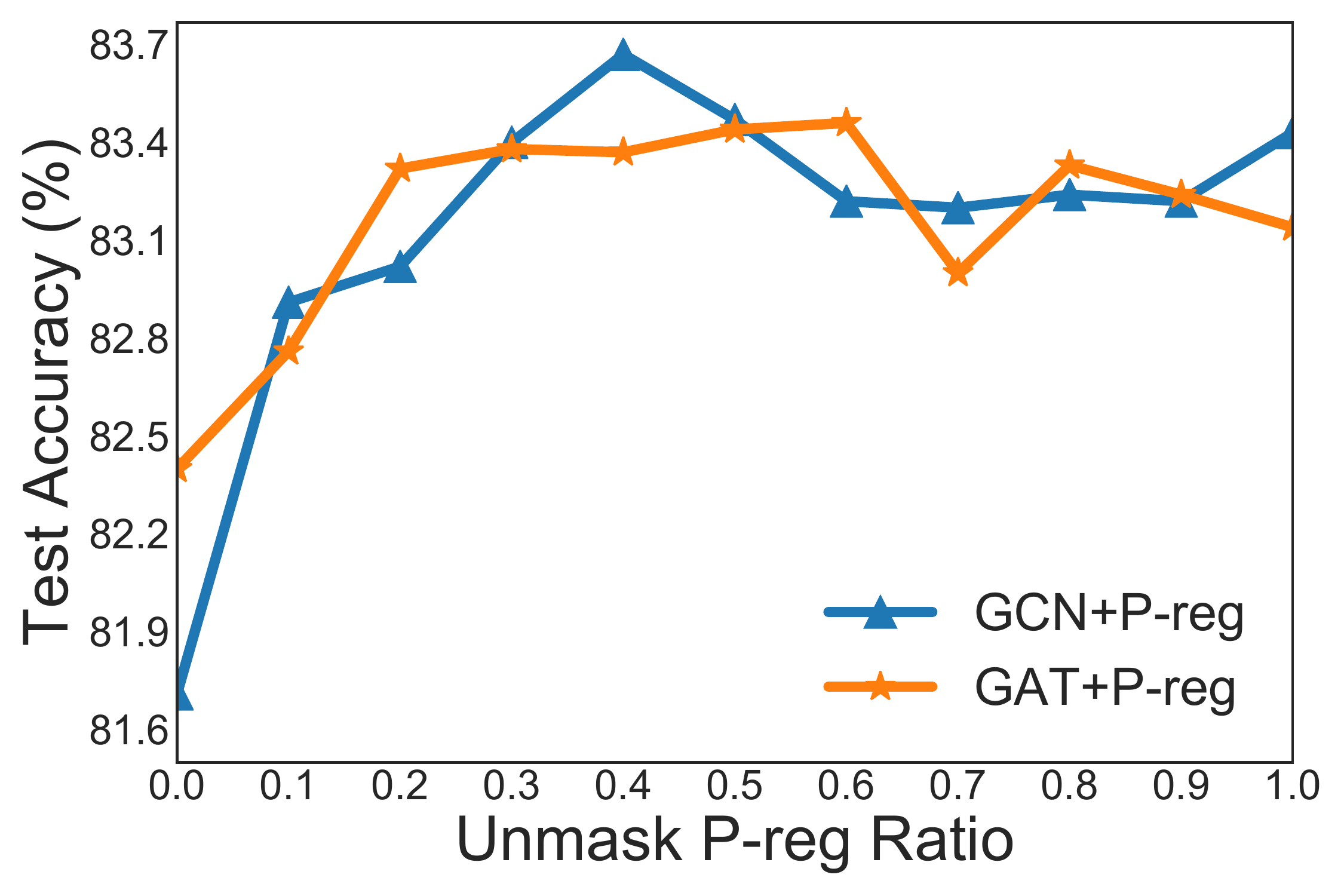}
		\caption{}
		\label{fig:mask-preg}
	\end{subfigure}
	\begin{subfigure}{0.33\textwidth}
		\includegraphics[width=\linewidth]{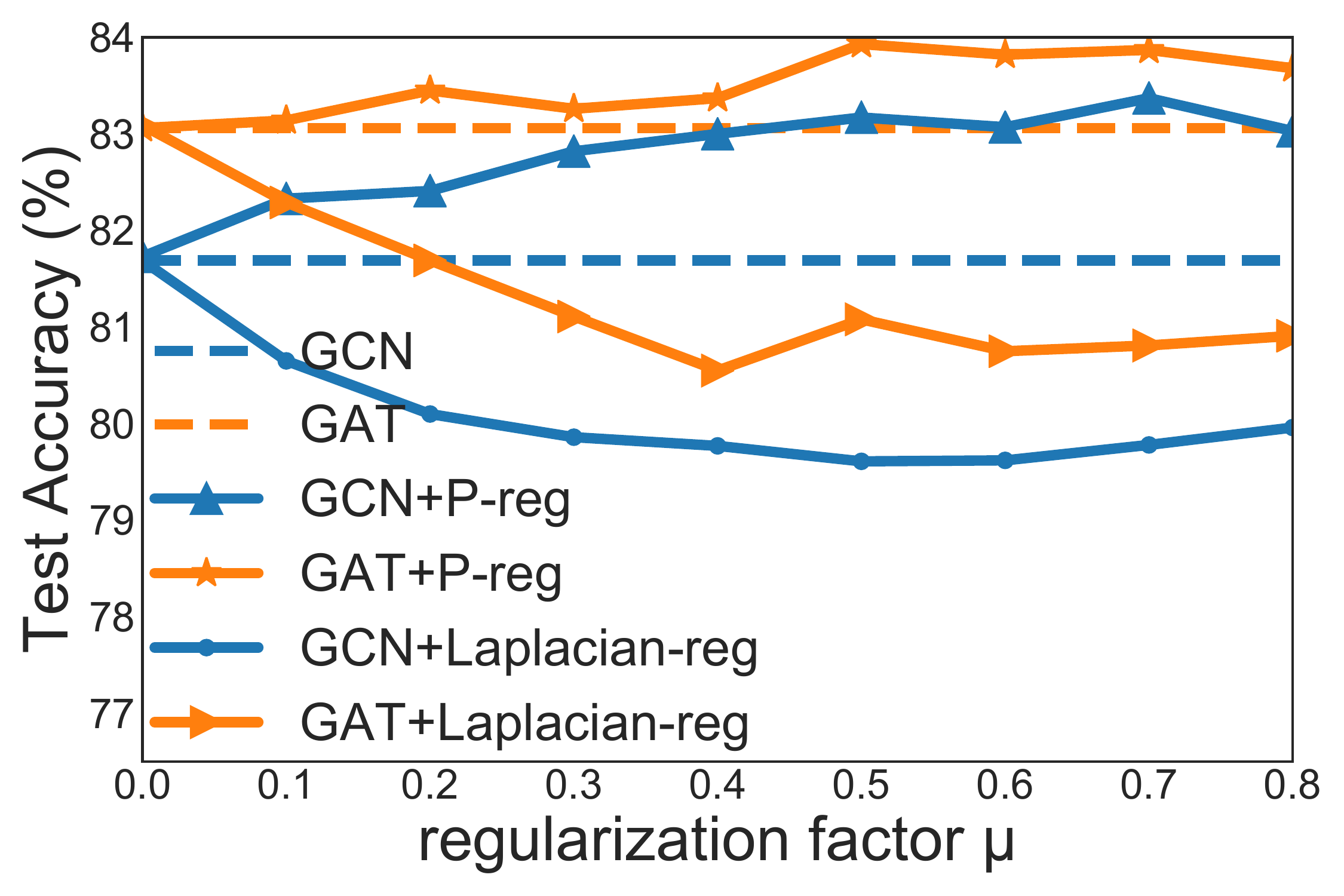}
		\caption{}
		\label{fig:preg-laplacian}
	\end{subfigure}
	\caption{The effects of increasing \textit{(a) the number of training nodes, (b) the number of nodes applied with P-reg, (c) regularization factor on the accuracy of GCN and GAT,} on the CORA dataset}
\end{figure*}
\begin{figure*}[t]
	\centering
	\begin{subfigure}{.32\textwidth}
		\centering
		\includegraphics[width=\linewidth]{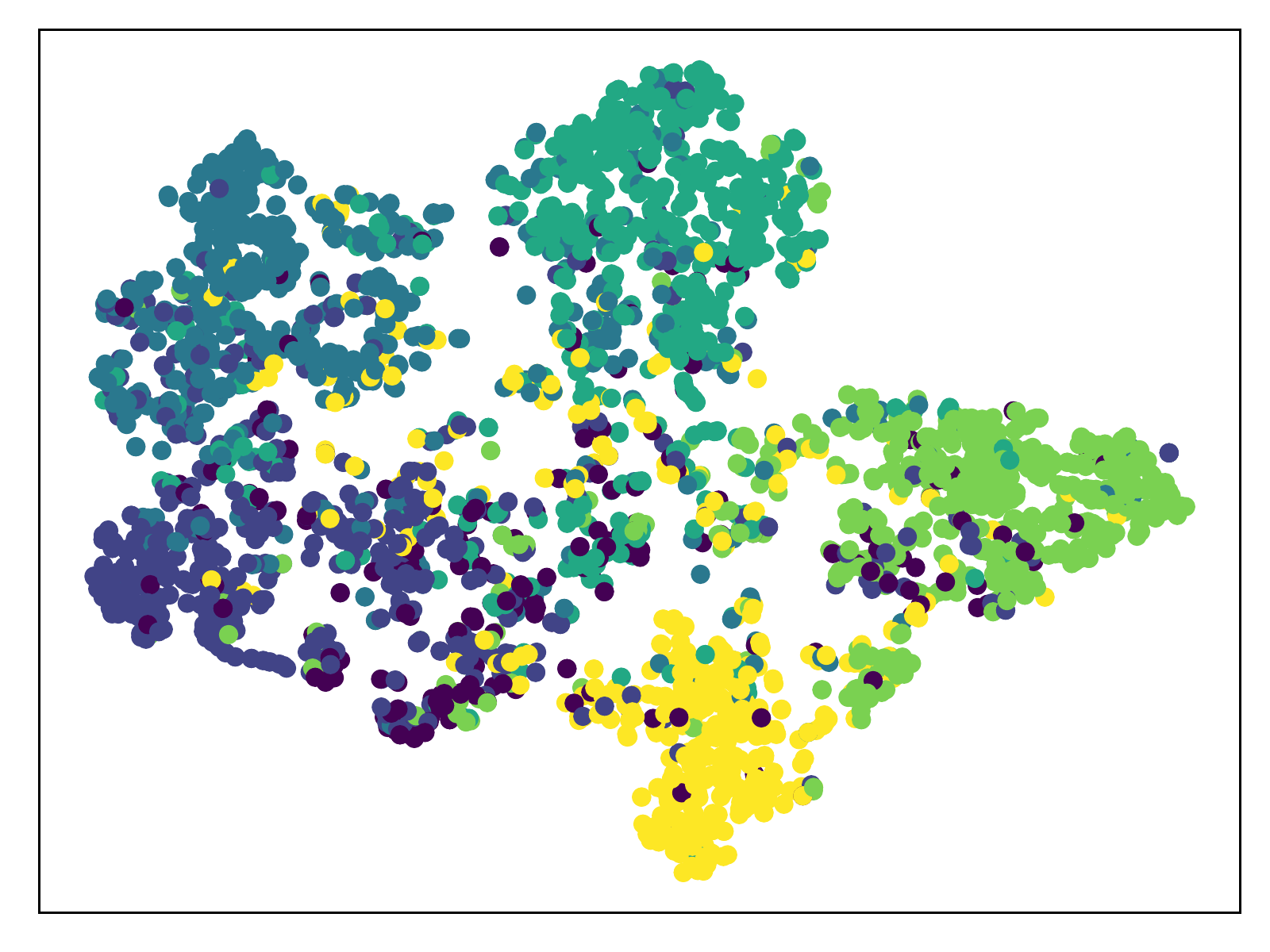}
		\caption{GCN}
		\label{fig:tsne1}
	\end{subfigure}%
	\begin{subfigure}{.32\textwidth}
		\centering
		\includegraphics[width=\linewidth]{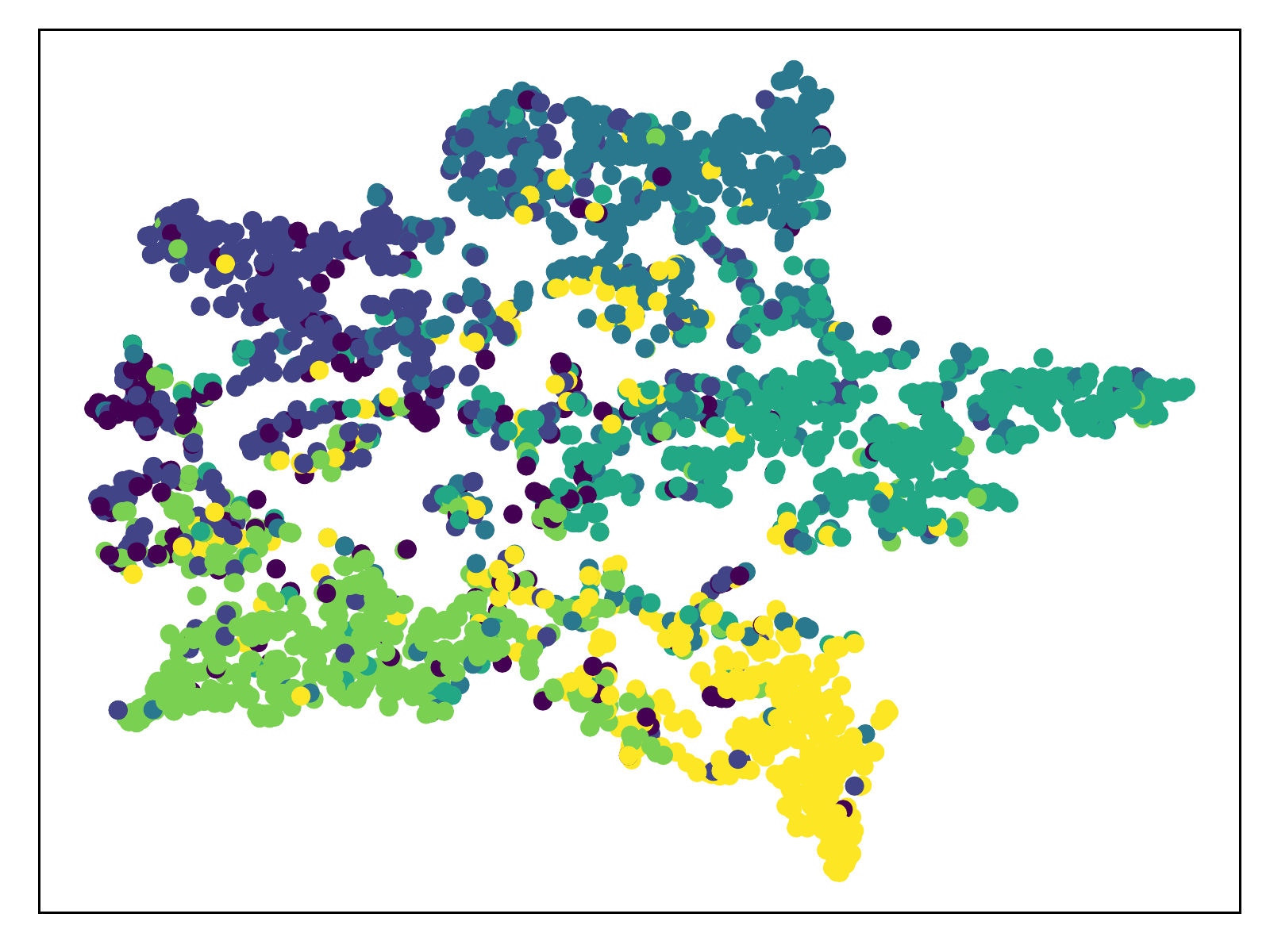}
		\caption{GCN+Laplacian-reg (\(\mu=0.5\))}
		\label{fig:tsne2}
	\end{subfigure}
	\begin{subfigure}{.32\textwidth}
		\centering
		\includegraphics[width=\linewidth]{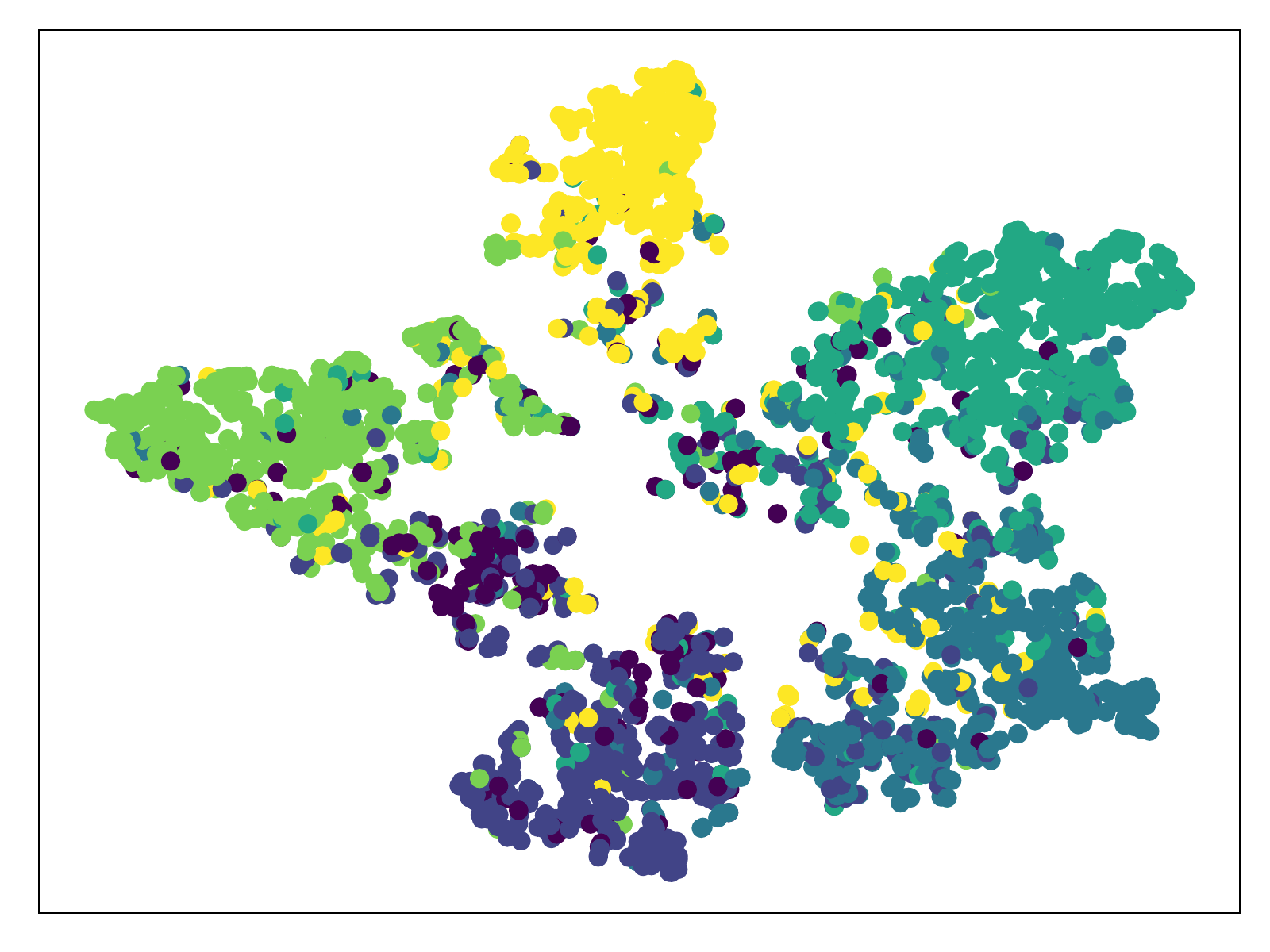}
		\caption{GCN+P-reg (\(\mu=0.5\))}
		\label{fig:tsne3}
	\end{subfigure}
	\caption{The t-SNE visualization of GCN outputs on the CiteSeer dataset (best viewed in color)}
	\label{fig:tsne}
\end{figure*}

\subsubsection{4.1.1 Graph Regularization Provides Extra Supervision Information}

\noindent \\
Usually, the more training nodes are provided, the more knowledge a model can learn and thus the higher accuracy the model can achieve. This can be verified in Figure~\ref{fig:training-nodes}:  as we keep adding nodes randomly to the training set, we generally obtain higher accuracy for the GCN/GAT models. 

However, in real-world graphs, the training nodes are usually only a small fraction of all nodes. Graph regularization, as an unsupervised term applied on a model's output in training, can provide extra supervision information for the model to learn a better representation for the nodes in a graph. Especially in semi-supervised learning, regularization can bridge the gap between labeled data and unlabeled data, making the supervised signal spread over the whole dataset. The effectiveness of regularization in semi-supervised learning has been shown in~\cite{zhu2003semi,zhou2003learning,belkin2006manifold,ando2007learning}.

\subsubsection{4.1.2 Limitations of Graph Laplacian Regularization}

\noindent \\ Unfortunately, \textbf{graph Laplacian regularization can hardly provide extra information that existing GNNs cannot capture.} The original graph convolution operator \(g_\theta\) is approximated by the truncated Chebyshev polynomials as \(g_\theta \ast Z = \sum_{k=0}^K \theta_k T_k(\tilde{\Delta})Z\)~\cite{hammond2011wavelets} in GNNs, where \(K\) means to keep \(K\)-th order of the Chebyshev polynomials, and \(\theta_k\) is the parameters of \(g_\theta\) as well as the Chebyshev coefficients. \(T_k(x)=2xT_{k-1}(x)-T_{k-2}(x)\)  is the Chebyshev polynomials, with \(T_0(x)=1\) and \(T_1(x) =x\).   \(K\) is equal to \(1\) for today's spatial GNNs such as GCN~\cite{kipf2017semi}, GAT~\cite{velickovic2018graph}, GraphSAGE~\cite{hamilton2017inductive} and so on. Thus, the 1-order Laplacian information is captured by popular GNNs, making the Laplacian regularization useless to them. This is also verified in Figure~\ref{fig:preg-laplacian}, which shows that the Laplacian regularization does not improve the test accuracy of GCN and GAT.

\subsubsection{4.1.3 How P-Reg Addresses the Limitations of Graph Laplacian Regularization}
\label{sec:preg-supervision}

\noindent \\ \textbf{P-reg provides new supervision information for GNNs.} Although P-reg shares similarities with Laplacian regularization, one key difference is that the Laplacian regularization \(\mathcal{L}_{lap} = \sum_{(i,j) \in \mathcal{E}} \left\| Z_i^\top - Z_j^\top\right\|_2^2 \) is \textit{edge-centric}, while P-reg \(\mathcal{L}_{P\text{-}reg}= \phi (Z, \hat{A}Z) = \sum_{i=1}^N \phi (Z_i^\top, (\hat{A}Z)_i^\top)\) is \textit{node-centric}.  The node-centric P-reg can be regarded as using the aggregated predictions of neighbors as the supervision target for each node. Based on the model's prediction vectors \(Z \in \mathbb{R}^{N \times C}\), the average of the neighbors' predictions of node \(v_i\), i.e., \((\hat{A}Z)_i^\top=\sum_{k \in \mathcal{N}(v_i)}z_k\), serves as the soft supervision targets of the model's output \(z_i\). This is similar to taking the \textit{voting result} of each node's neighbors to supervise the nodes. Thus, P-reg provides additional categorical information for the nodes, which cannot be obtained from Laplacian regularization as it simply pulls the representation of edge-connected nodes closer.

\begin{figure*}[t]
	\centering
	\begin{subfigure}{.33\textwidth}
		\centering
		\includegraphics[width=\linewidth]{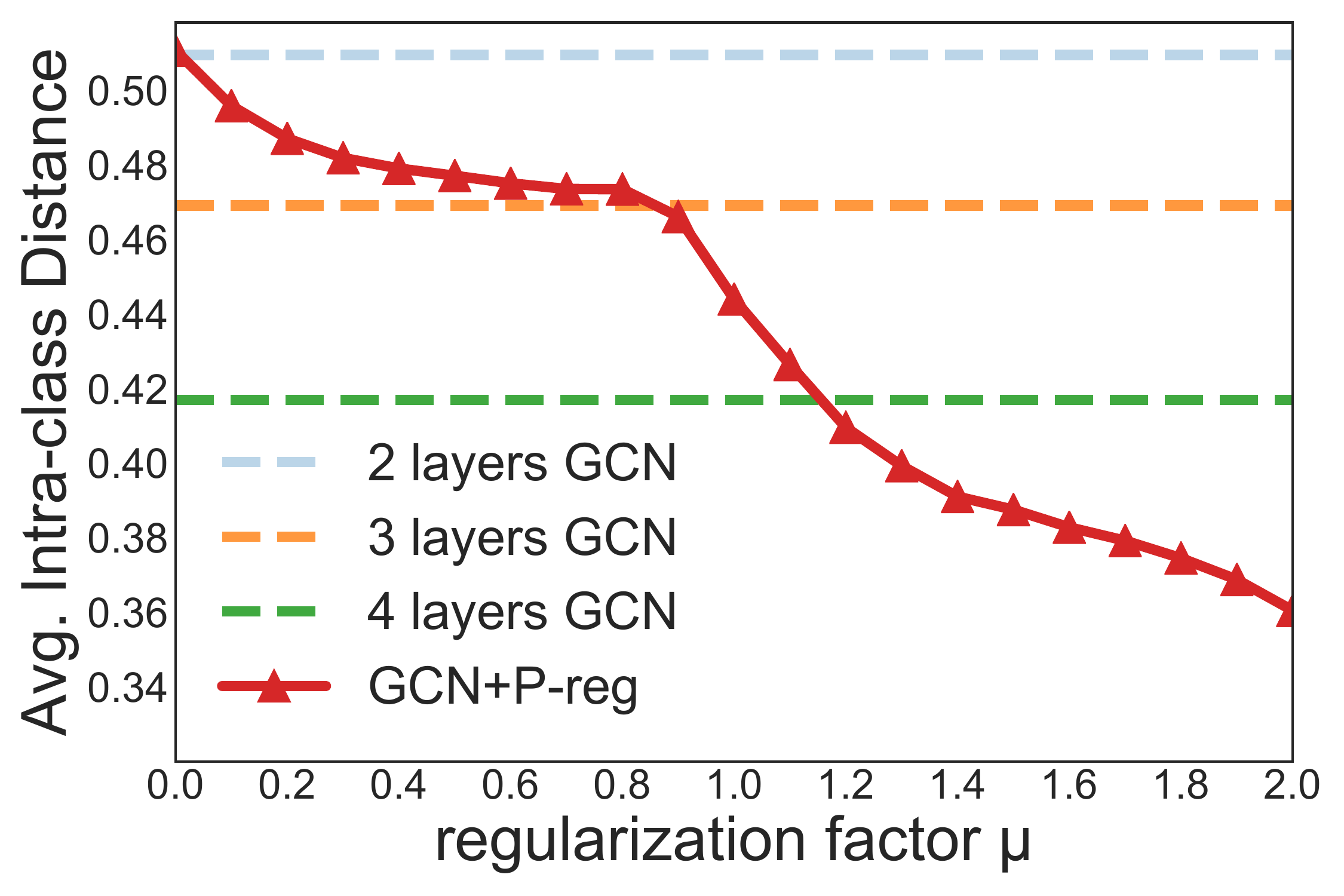}
		\caption{Avg. intra-class distance \(\omega\)}
		\label{fig:idgcn-1}
	\end{subfigure}%
	\begin{subfigure}{.33\textwidth}
		\centering
		\includegraphics[width=\linewidth]{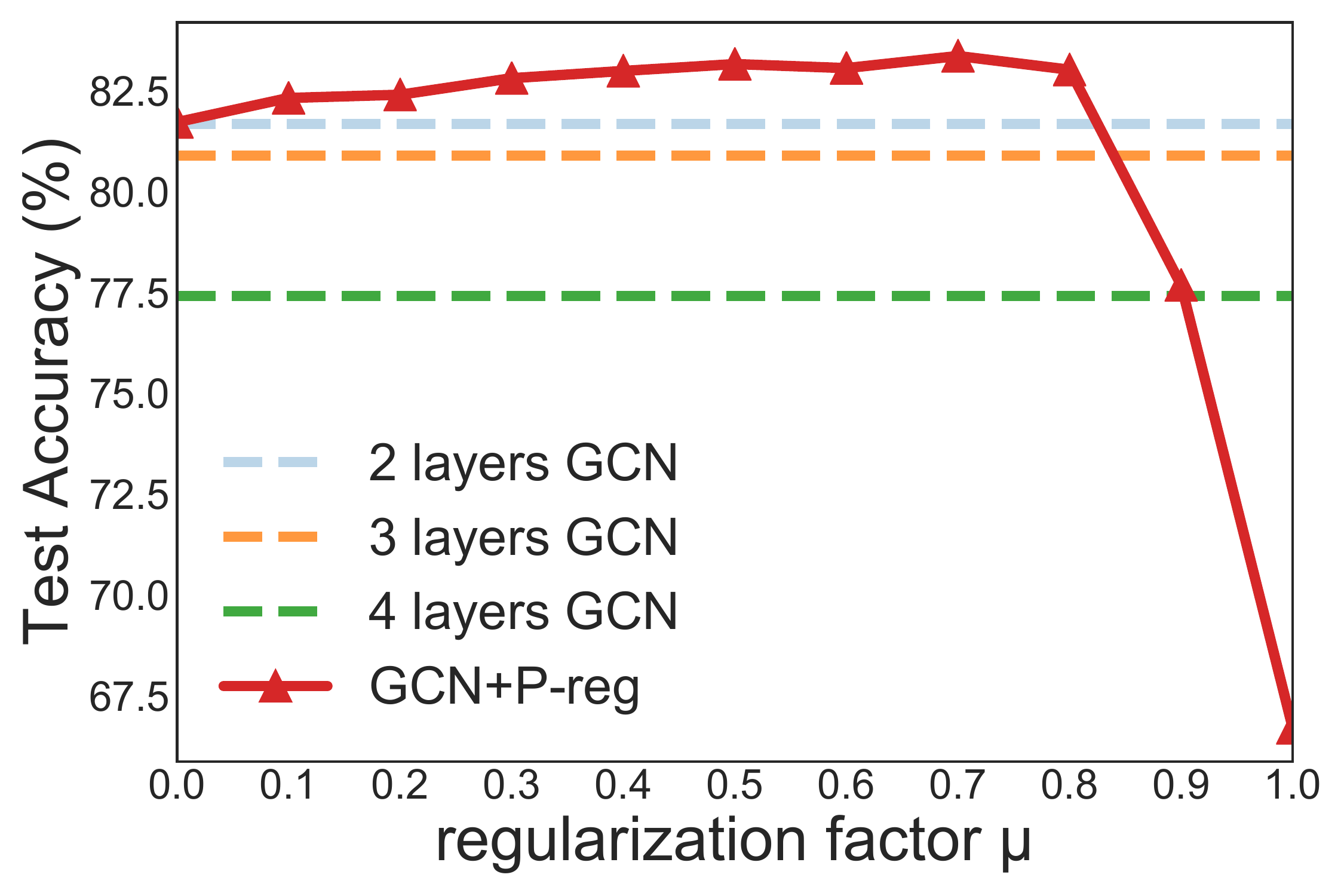}
		\caption{Test accuracy}
		\label{fig:idgcn-2}
	\end{subfigure}
	\begin{subfigure}{.33\textwidth}
		\centering
		\includegraphics[width=\linewidth]{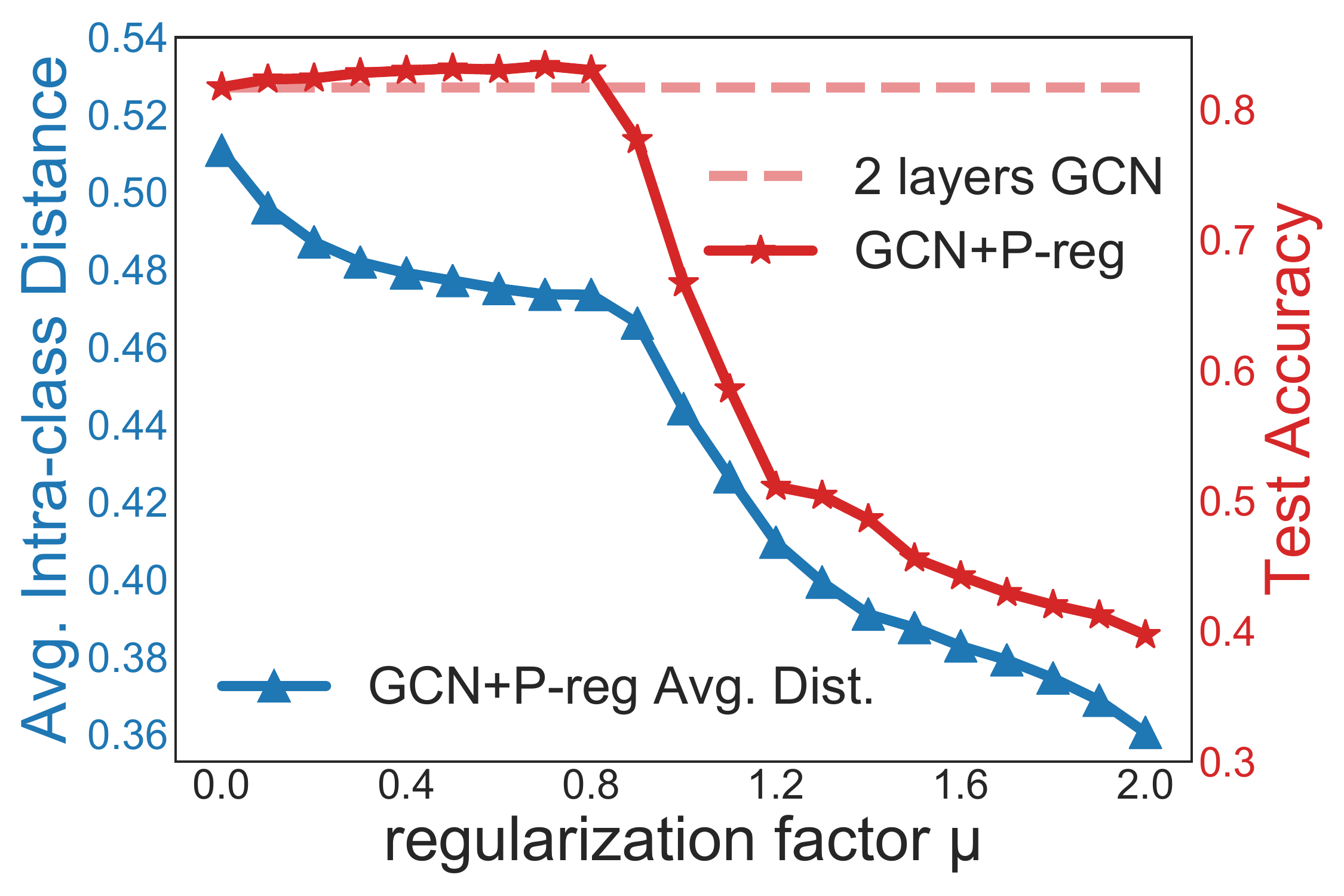}
		\caption{Relation between \(\omega\) and accuracy}
		\label{fig:idgcn-3}
	\end{subfigure}
	\caption{Empirical observation of connections between P-reg and deep GCNs (best viewed in color)}
	\label{fig:idgcn}
\end{figure*}

\noindent \textbf{Empirical verification.} The above analysis can be verified from the node classification accuracy in Figure~\ref{fig:mask-preg}. As P-reg is node-centric, we can add a mask to a node $v_i$ so that unmasking $v_i$ means to apply \(\phi\) to $v_i$. Define the unmask ratio as \(\alpha = \frac{|S_{\text{unmask}}|}{N}\), where \(S_{\text{unmask}}\) is the set of nodes with \(\phi\) applied. The masked P-reg can be defined as \(\mathcal{L}_{P\text{-}reg\text{-}m}=\frac{1}{|S_{\text{unmask}}|}\sum_{i \in S_{\text{unmask}}}\phi(Z_i^\top, (AZ)_i^\top)\). If the unmask ratio \(\alpha\) is \(0\), the model is basically vanilla GCN/GAT. As the unmask ratio increases, P-reg is applied on more nodes and Figure~\ref{fig:mask-preg} shows that the accuracy generally increases. After sufficient nodes are affected by P-reg, the accuracy tends to be stabilized. The same result is also observed on all the datasets in our experiments. The result thus verifies that we can consider \(\hat{A}Z\) as the supervision signal of \(Z\), and the more supervision we have, the higher is the accuracy. %

In Figure~\ref{fig:preg-laplacian}, we show that P-reg improves the test accuracy of both GCN and GAT, while applying Laplacian regularization even harms their performance. We also plot the t-SNE visualization in Figure~\ref{fig:tsne}. Compared with pure GCN and GCN with Laplacian regularization, Figure~\ref{fig:tsne3} shows that by applying P-reg, the output vectors of GCN are more condensed within each class and the classes are also more clearly separated from each other.

\subsection{Benefits of P-Reg from the Deep GCN Perspective}
\label{sec:preg-deepgcn}
\subsubsection{4.2.1 Deeper GCN Provides Information from Farther Nodes}

\noindent \\  \citet{xu2018representation} proved that for a \(K\)-layer GCN, the influence distribution\footnote{The influence distribution \(I_x\) is defined as the normalized influence score \(I(x,y)\), where the influence score \(I(x, y)\) is the sum of \(\begin{bmatrix}
\nicefrac{\delta h_x^{(k)}}{\delta h_y^{(0)}} 
\end{bmatrix}\), where \(h_x^{(k)}\) is the \(k\)-th layer activation of node \(x\). }  \(I_x\) for any node \(x\) in a graph $G$ is equivalent to the \(k\)-step random walk distribution on \(G\) starting at node \(x\). Thus, the more layers a GCN has, the more information a node \(x\) can obtain from nodes farther away from \(x\) in the graph $G$. Some works such as~\cite{li2019deepgcns,rong2020dropedge,huang2020tackling,zhao2020pairnorm} provide methods to make GNN models deep.

\subsubsection{4.2.2 Limitations of Deep GCNs}
\noindent \textbf{\\More layers is not always better.} The more layers a GCN has, the representation vectors of nodes produced become more prone to be similar. As shown in Lemma~\ref{lemma:idgcn}, an infinite-depth GCN even gives the same output vectors for all nodes. This is known as the over-smoothing problem of GNNs and there are studies~\citep{nt2019revisiting,oono2020graph} focusing on this issue, which claim that a GCN can be considered as a low-passing filter and exponentially loses its expressive power as the number of layers increases. Thus, more layers may make the representation vectors of nodes in-discriminable and thus can be detrimental to the task of node classification.

\noindent \textbf{Computational issues.} %
First, generally for all deep neural networks, gradient vanishing\-/exploding issues will appear when the neural networks go deeper. Although it could be mitigated by using residual skip connections, the training of deeper neural networks will inevitably become both harder and slower. Second, as GNNs grow deeper, the size of a node's neighborhood expands exponentially~\cite{huang2018adaptive}, making the computational cost for both gradient back-propagation training and feed-forwarding inference extremely high.

\subsubsection{4.2.3 How P-Reg Addresses the Limitations of Deep GCNs} 

\noindent \\ \textbf{P-reg can effectively balance information capturing and over-smoothing at low cost.} Although an infinite-depth model outputting the same vector for each node is not desired, only being able to choose the number of layers \(L\) from a discrete set \(\mathbb{N}_+\) is also not ideal. Instead, a balance between the size of the receptive field (the farthest nodes that a node can obtain information from) and the preservation of discriminating information is critical for generating useful node representation. To this end,  the regularization factor~\(\mu \in \mathbb{R}\) in Eq.~(\ref{eq:preg-gnn}) plays a vital role in enforcing a flexible strength of regularization on the GNN. When \(\mu=0\), the GNN model is basically the vanilla model without P-reg. On the contrary, the model becomes an infinite-depth GCN when \(\mu \rightarrow +\infty\). Consequently, using P-reg with a \textit{continuous} \(\mu\) can be regarded as using a GCN with a \textit{continuous} rather than discrete number of layers. Adjusting \(\mu\) empowers a GNN model to balance between information capturing and over-smoothing.

P-reg provides a controllable receptive field for a GNN model and this is achieved at only a computation cost of \(\mathcal{O}(NC+\left|\mathcal{E}\right|C)\), which is basically equivalent to the overhead of just adding one more layer of GCN\footnote{One layer of GCN has a computation cost of \(\mathcal{O}(\left|\mathcal{E}\right|C)\). Since \(2|\mathcal{E}| \geq N\) holds for all undirected connected graphs, we have \(\mathcal{O}(NC+\left|\mathcal{E}\right|C) = \mathcal{O}(3\left|\mathcal{E}\right|C) = \mathcal{O}(\left|\mathcal{E}\right|C)\).}. We also do not need to worry about gradient vanishing/exploding and neighborhood size explosion. P-reg also has zero overhead for GNN inference.

\begin{table*}[t]
	\caption{Accuracy improvements brought by P-reg and other techniques using \textbf{random splits}}
	\label{table:exp-results}
	\resizebox{\linewidth}{!}{%
		\begin{tabular}{@{}cllllllll@{}}
			\toprule
			&                       & \multicolumn{1}{c}{CORA} & \multicolumn{1}{c}{CiteSeer} & \multicolumn{1}{c}{PubMed} & \multicolumn{1}{c}{CS} & \multicolumn{1}{c}{Physics} & \multicolumn{1}{c}{Computers} & \multicolumn{1}{c}{Photo} \\ \midrule
			\multirow{5}{*}{GCN} & Vanilla   & 79.47\PM{0.89} & 69.45\PM{0.99} & 76.26\PM{1.30} & 91.48\PM{0.42} & 93.66\PM{0.41} & 80.73\PM{1.52} & 89.24\PM{0.76} \\
			\cmidrule(l){2-9}
			& Label Smoothing & 79.70\PM{1.23} & 70.08\PM{1.48} & 76.83\PM{1.20} & 91.65\PM{0.49} & 93.67\PM{0.41} & 81.31\PM{1.46} & 90.03\PM{0.81} \\
			& Confidence Penalty & 79.70\PM{1.08} & 69.69\PM{1.22} & 76.57\PM{0.97} & 91.55\PM{0.45} & 93.67\PM{0.41} & 80.70\PM{1.50} & 89.26\PM{0.76} \\
			& Laplacian Regularizer  & 79.41\PM{0.65}  & 69.46\PM{1.03}  & 76.29\PM{1.13}  & 91.51\PM{0.42}  & 93.65\PM{0.40}  & 81.20\PM{1.69}  & 89.52\PM{0.57} \\
			\cmidrule(l){2-9} 
			& P-reg   & \textbf{82.83}\PM{1.16}  & \textbf{71.63}\PM{2.17}  & \textbf{77.37}\PM{1.50}  & \textbf{92.58}\PM{0.32}  & \textbf{94.41}\PM{0.74}  & \textbf{81.66}\PM{1.42}  & \textbf{91.24}\PM{0.75} \\
			\midrule
			\multirow{5}{*}{GAT} & Vanilla & 79.47\PM{1.77} & 69.28\PM{1.56} & 75.61\PM{1.56} & 91.15\PM{0.36} & 93.08\PM{0.55} & 81.00\PM{2.02} & 89.55\PM{1.20} \\
			\cmidrule(l){2-9}
			& Label Smoothing   & 80.30\PM{1.71}  & 69.42\PM{1.41}  & 76.53\PM{1.28}  & 91.09\PM{0.34}  & 93.25\PM{0.51}  & 81.92\PM{1.98}  & 90.55\PM{0.86} \\
			& Confidence Penalty   & 79.89\PM{1.73}  & 69.87\PM{1.35}  & 76.44\PM{1.53}  & 90.91\PM{0.36}  & 93.13\PM{0.53}  & 78.86\PM{2.07}  & 88.58\PM{1.96} \\
			& Laplacian Regularizer  & 80.23\PM{1.90}  & 69.50\PM{1.42}  & \textbf{76.80}\PM{1.57}  & 90.86\PM{0.39}  & 93.09\PM{0.52}  & 82.38\PM{2.00}  & 90.58\PM{1.15} \\
			\cmidrule(l){2-9}
			& P-reg     & \textbf{82.97}\PM{1.19}  & \textbf{70.00}\PM{1.89}  & 76.39\PM{1.46}  & \textbf{91.92}\PM{0.20}  & \textbf{94.28}\PM{0.29}  & \textbf{83.68}\PM{2.24}  & \textbf{91.31}\PM{1.06} \\
			\midrule
			\multirow{5}{*}{MLP} & Vanilla & 57.47\PM{2.46} & 56.96\PM{2.10} & 68.36\PM{1.35} & 86.87\PM{1.01} & 89.43\PM{0.67} & 62.61\PM{1.81} & 76.26\PM{1.40} \\
			\cmidrule(l){2-9}
			& Label Smoothing   & 59.00\PM{1.54}  & 57.98\PM{1.76}  & 68.75\PM{1.26}  & 87.90\PM{0.57}  & 89.54\PM{0.61}  & 62.55\PM{2.21}  & 76.12\PM{1.11} \\
			& Confidence Penalty  & 57.20\PM{2.59}  & 56.89\PM{2.15}  & 68.16\PM{1.36}  & 86.97\PM{1.04}  & 89.54\PM{0.63}  & 62.75\PM{2.15}  & 76.19\PM{1.34} \\
			& Laplacian Regularizer  & 60.30\PM{2.47}  & 58.62\PM{2.40}  & 68.67\PM{1.39}  & 86.96\PM{0.75}  & 89.50\PM{0.48}  & 62.60\PM{1.99}  & 76.23\PM{1.09} \\
			\cmidrule(l){2-9}
			& P-reg  & \textbf{64.41}\PM{4.56}  & \textbf{61.09}\PM{2.13}  & \textbf{70.09}\PM{1.75}  & \textbf{90.87}\PM{1.90}  & \textbf{91.57}\PM{0.69}  & \textbf{68.93}\PM{3.28}  & \textbf{79.71}\PM{3.72} \\
			\bottomrule
		\end{tabular}
	}
\end{table*}

\begin{table}[t]
	\caption{Comparison of P-reg with the state-of-the-art methods using the \textbf{standard split}}
	\label{table:sota}
	\centering
	{ %
	\begin{tabular}{lccc}
	  \toprule
	  & {CORA} & {CiteSeer} & {PubMed} \\
	  \midrule
	  GCN & 81.67\PM{0.60} & 71.57\PM{0.46} & 79.17\PM{0.37} \\
	  GAT & 83.20\PM{0.72} & 71.29\PM{0.84} & 77.99\PM{0.47}\\
	  APPNP  & 83.12\PM{0.44} & 72.00\PM{0.48} & 80.10\PM{0.26} \\
	  GMNN & 83.18\PM{0.85} & 72.77\PM{1.43} & \textbf{81.63}\PM{0.36} \\
	  Graph U-Nets & 81.31\PM{1.47} & 67.74\PM{1.48} &77.76\PM{0.84} \\
	  GraphAT & 82.46\PM{0.60} & 73.49\PM{0.38} & 79.10\PM{0.20}\\
	  BVAT  & 83.37\PM{1.01} & 73.83\PM{0.54} & 77.97\PM{0.84}\\
	  GraphMix  & 83.54\PM{0.72} & 73.81\PM{0.85} & 80.76\PM{0.84} \\
	  GAM  & 82.28\PM{0.48} & 72.74\PM{0.62} & 79.60\PM{0.63} \\
	  DeepAPPNP & 83.78\PM{0.33} & 71.37\PM{0.66} & 79.73\PM{0.22} \\
	  \midrule
	  GCN+P-reg & 83.38\PM{0.86} & \textbf{74.83}\PM{0.17} & 80.11\PM{0.45} \\
	  GAT+P-reg & \textbf{83.89}\PM{0.31} & 72.86\PM{0.65} & 78.39\PM{0.27}\\
	  \bottomrule
	\end{tabular}
  }
  \end{table}

\noindent \textbf{Empirical verification.} Define \(\omega = \frac{1}{N}\sum_{k=1}^C \sum_{i \in S_k } \|z_i - c_k\|_2\) as the average intra-class Euclidean distance for a trained GNN's output \(Z\), where \(z_i\) is the learned representation of node \(v_i\) (the \(i\)-th row of \(Z\)),  \(S_k\) denotes all nodes of class \(k\) in the graph, and \(c_k = \frac{1}{|S_k|}\sum_{i \in S_k}z_i\). We trained GCNs with 2, 3 and 4 layers and GCNs+P-reg with different \(\mu\). The results on the CORA graph are reported in Figure~\ref{fig:idgcn}, where all y-axis values are the average values of 10 random trials. Similar patterns are also observed on other datasets in our experiments.

Figure~\ref{fig:idgcn-1} shows that as a GCN has more layers, \(\omega\) becomes smaller  (as indicated by the 3 dashed horizontal lines). Also, as \(\mu\) increases,  \(\omega\) for GCN+P-reg becomes smaller. The result is consistent with our analysis that a larger \(\mu\) corresponds to a deeper GCN.  Figure~\ref{fig:idgcn-2} further shows that as \(\mu\) increases, the test accuracy of GCN+P-reg  first improves and then becomes worse. The best accuracy is achieved when \(\mu=0.7\). By checking Figure~\ref{fig:idgcn-1}, \(\mu=0.7\) corresponds to a point lying between the lines of the 2-layer and 3-layer GCNs. This implies that the best accuracy could be achieved with an $l$-layer GCN, where \(2<l<3\). While a \textit{real} $l$-layer GCN does not exist, applying P-reg to GCN produces the equivalent result as from an $l$-layer GCN. Finally, Figure~\ref{fig:idgcn-3} shows the relation between \(\omega\) and the accuracy. The high accuracy region corresponds to the flat area of \(\omega\). This could be explained as the flat area of \(\omega\) achieves a good balance between information capturing and over-smoothing.

\section{Experimental Results}

\begin{table*}[t]
	\caption{The scores (with standard deviation) of graph-level tasks of P-reg applied on GCN/GIN on the OGB datasets (ROC-AUC scores: higher is better. RMSE scores: lower is better.)}
	\label{table:graph-class}
	\centering
	\begin{tabular}{lccccc}
	  \toprule
						& moltox21  & molhiv    & molbbbbp  & molesol & molfreesolv \\
						& (ROC-AUC)$\uparrow$ & (ROC-AUC)$\uparrow$ & (ROC-AUC)$\uparrow$ & (RMSE)$\downarrow$ & (RMSE)$\downarrow$ \\
	  \midrule
	  GCN               &	74.75\PM{0.55} & \textbf{76.08}\PM{1.41}	& 67.99\PM{1.18} & 1.124\PM{0.034} & 2.564\PM{0.148}	\\
	  GCN+P-reg         & 75.86\PM{0.75} & 76.03\PM{1.25} & 69.37\PM{1.03} & 1.141\PM{0.038} & 2.406\PM{0.147} \\
	  GCN+Virtual       & 77.42\PM{0.67} & 75.67\PM{1.54} & 66.92\PM{0.85} & 1.020\PM{0.056} & 2.164\PM{0.129} \\
	  GCN+Virtual+P-reg & \textbf{77.58}\PM{0.39} & 75.81\PM{1.61} & \textbf{69.67}\PM{1.62} & \textbf{0.946}\PM{0.041} & \textbf{2.082}\PM{0.119} \\
	  \midrule
	  GIN               & 74.89\PM{0.66} & 75.47\PM{1.22} & 68.03\PM{2.72} & 1.150\PM{0.057} & 2.971\PM{0.263} \\
	  GIN+P-reg         & 74.90\PM{0.59} & 76.77\PM{1.41} & 67.69\PM{2.43} & 1.137\PM{0.028} & 2.529\PM{0.216} \\
	  GIN+Virtual       & 77.04\PM{0.83} & 76.33\PM{1.52} & 68.37\PM{2.08} & 0.991\PM{0.052} & 2.172\PM{0.192} \\
	  GIN+Virtual+P-reg & \textbf{77.51}\PM{0.68} & \textbf{77.12}\PM{1.14} & \textbf{70.25}\PM{1.86} & \textbf{0.965}\PM{0.069} & \textbf{2.050}\PM{0.144} \\
	  \bottomrule
	\end{tabular}
  \end{table*}
We evaluated P-reg on node classification, graph classification and graph regression tasks. We report the results of P-reg here using Cross Entropy as  \(\phi\) if not specified, and the results of different choices of the \(\phi\) function are reported in Appendix~F due to the limited space. 
In addition to the node classification task for both random splits of 7 graph datasets~\cite{yang2016revisiting,mcauley2015image,shchur2019pitfalls} and the standard split of 3 graph datasets reported in Section~\ref{exp:random_split} and \ref{exp:standard_split}, we also evaluated P-reg on graph-level tasks on the OGB dataset~\cite{hu2020open} in Section~\ref{exp:graph-level}.
Our implementation is based on PyTorch~\cite{pytorch} and we used the Adam~\cite{kingma2014adam} optimizer with  learning rate equal to \(0.01\) to train all the models. 
Additional details about the experiment setup are given in Appendix~E.

\subsection{Improvements on Node Classification Accuracy}
\label{exp:random_split}

We evaluated the node classification accuracy for 3 models on 7 popular datasets using random splits.  The train\-/validation\-/test  split of all the 7 datasets are 20 nodes\-/30 nodes\-/all the remaining nodes per class, as recommended by~\citet{shchur2019pitfalls}. We conducted each experiment on 5 random splits and 5 different trials for each random split. The mean value and standard deviation are reported in Table~\ref{table:exp-results} based on the \(5\times 5\) experiments for each cell. The regularization factor \(\mu\) is determined by grid search using the validation accuracy. The search space for \(\mu\) is 20 values evenly chosen from \(\left[0, 1\right]\). The \(\mu\) is the same for each cell (model \(\times\) dataset) in  Table~\ref{table:exp-results}.

We applied P-reg on GCN,
GAT and MLP, respectively. GCN is a typical convolution-based GNN, while GAT is attention based. Thus, they are representative of a broad range of GNNs. MLP is multi-layer perception without modeling the graph structure internally. In addition, we also applied label smoothing, confidence penalty, and Laplacian regularization to the models, respectively, in order to give a comparative analysis on the effectiveness of P-reg. Label smoothing and confidence penalty are two general techniques used to improve the generalization capability of a model.  Label smoothing~\citep{szegedy2016rethinking,rafael2019when} has been adopted in many state-of-the-art deep learning models such as~\cite{huang2019gpipe,real2019regularized,vaswani2017attention,zoph2018learning} in computer vision and natural language processing. It softens the one-hot hard targets \(y_c=1,\ y_i=0 \ \forall i \neq c\) into \(y_i^{LS} = \left(1-\alpha\right)y_i + \nicefrac{\alpha}{C}\), where \(c\) is the correct label and \(C\) is the number of classes.  Label smoothing 
Confidence penalty~\cite{pereyra2017regularizing} adds the negative entropy of the network outputs to the classification loss as a regularizer, \(\mathcal{L}^{CP}=\mathcal{L}_{cls} + \beta \sum_{i} p_i \log p_i\), where \(p_i\) is the predicted class probability of the \(i\)-th sample.

Table~\ref{table:exp-results} shows that P-reg significantly improves the accuracy of both GCN and GAT, as well as MLP. General techniques such as label smoothing and confidence penalty also show their capability to improve the accuracy on most datasets but the improvements are relatively small compared with the improvements brought by P-reg. The improvements by P-reg are also consistent in all the cases except for GAT on the PubMed dataset. Although Laplacian regularization improves the performance of MLP on most datasets, it  has marginal improvement or even negative effects on GCN and GAT, which also further validates our analysis in Section~\ref{why:reg}.

\subsection{Comparison with the State-of-the-Art Methods (using the Standard Split)}\label{exp:standard_split}

We further compared GCN+P-reg and GAT+P-reg with the state-of-the-art methods.
Among them, APPNP~\cite{klicpera2018predict}, GMNN~\cite{qu2019gmnn} and Graph U-Nets\cite{gao2019graph} are newly proposed state-of-the-art GNN models, and GraphAT~\citep{feng2019graph}, BVAT~\citep{deng2019batch} and GraphMix~\citep{verma2019graphmix} use various complicated techniques to improve the performance of GNNs. GraphAT~\citep{feng2019graph} and BVAT~\citep{deng2019batch} incorporate the  adversarial perturbation\footnote{Adversarial perturbation is the perturbation along the direction of the model gradient \(\frac{\partial f}{\partial x}\).} into the input data. GraphMix~\citep{verma2019graphmix} adopts the idea of co-training~\citep{blum1998combining} to use a parameters-shared fully-connected network to make a GNN more generalizable. It combines many other semi-supervised techniques such as Mixup~\citep{zhang2017mixup}, entropy minimization with Sharpening~\citep{gradvalet2005semi}, Exponential Moving Average of predictions~\citep{tarvainen2017mean} and so on. GAM~\citep{otilia2019graph} uses co-training of GCN with an additional agreement model that gives the probability that two nodes have the same label. The DeepAPPNP~\cite{rong2020dropedge, huang2020tackling} we used contains \(K\! =\!64\) layers with restart probability $\alpha\! =\! 0.1$ and DropEdge rate $0.2$. If more than one method/variant is proposed in the respective papers, we report the best performance we obtained for each work using their official code repository or PyTorch-geometric~\cite{pytorch-geometric} benchmarking code.

Table~\ref{table:sota} reports the test accuracy of node classification using the standard split, on the CORA, CiteSeer and PubMed datasets. We report the mean and standard deviation of the accuracy of 10 different trials. The search space for \(\mu\) is 20 values evenly chosen from \([0, 1]\). P-reg outperforms the state-of-the-art methods on  CORA and CiteSeer, while its performance is among the best three on PubMed. This result shows that P-reg's performance is very competitive because P-reg is a simple regularization with only a single hyper-parameter, unlike in the other methods where heavy tricks, adversarial techniques or complicated models are used. %

\subsection{Experiments of Graph-Level Tasks}
\label{exp:graph-level}
We also evaluated the performance of  P-reg on graph-level tasks on the OGB (Open Graph Benchmark)~\cite{hu2020open} dataset. We reported the mean and standard deviation of the scores for the graph-level tasks of 10 different trials in Table~\ref{table:graph-class}. The ROC-AUC scores of the graph classification task on the \textit{moltox21, molhiv, molbbbbp} datasets are the higher the better, while the RMSE scores of the graph regression task on the \textit{molesol, molfreesolv} datasets are the lower the better. We use the code of GCN, Virtual GCN, GIN, Virtual GIN provided in the OGB codebase.  Here \textit{Virtual} means that the graphs are augmented with virtual nodes~\cite{gilmer2017neural}. GIN~\cite{xu2018how} represents the state-of-the-art backbone GNN model for graph-level tasks.  P-reg was added directly to those models, without modifying any other configurations. The search space for \(\mu\) is 10 values evenly chosen from \([0, 1]\). %

Table~\ref{table:graph-class} shows that P-reg can  improve the performance of GCN and GIN, as well as Virtual GCN and Virtual GIN, in most cases for both the graph classification task and the graph regression task. In addition, even if \textit{virtual node}  already can effectively boost the performance of GNN models in the graph-level tasks, using P-reg can still further improve the performance. As shown in Table~\ref{table:graph-class}, Virtual+P-reg makes the most improvement over vanilla GNNs on most of the datasets.

\section{Conclusions}

We presented P-reg, which is a simple graph regularizer designed to boost the performance of existing GNN models. We theoretically established its connection to graph Laplacian regularization and its equivalence to an infinite-depth GCN. We also showed that P-reg can provide new supervision signals and simulate a deep GCN at low cost while avoiding over-smoothing. We then validated by experiments that compared with existing techniques, P-reg is a significantly more effective method that consistently improves the performance of popular GNN models such as GCN, GAT and GIN. %

\section{ Acknowledgments}
We thank the reviewers for their valuable comments. This work was partially supported GRF 14208318 from the RGC and ITF 6904945 from the ITC of HKSAR, and the National Natural Science Foundation of China (NSFC) (Grant No. 61672552).

\bibliography{main}

\newpage

\appendix
\onecolumn

{
\centering
{\LARGE\bf Supplementary Materials For ``Rethinking Graph Regularization for Graph Neural Networks''  \par}

}

\section{Notations}
\label{app:notations}
\begin{table}[h]\small
\caption{Notations}
\centering
\begin{tabular}{ll}
    \toprule
    \(G\) & a graph\\
\midrule
\(\mathcal{V}\) & the vertex set of a graph\\
\midrule
\(\mathcal{E}\) & the edge set of a graph\\
\midrule
\(G(A, X)\) & a graph G with its adjacency matrix \(A\) and node feature matrix \(X\)\\
\midrule
\(v_i\) & the \(i\)-th node in a graph\\
\midrule
\(\mathcal{N}(v_i)\) & the neighborhood of \(v_i\)\\
\midrule
\(A\) & \(A \in \mathbb{R}^{N \times N}\), the adjacency matrix of a graph\\
\midrule
\(D\) & the diagonal degree matrix, \(D_{ii} = \sum_{k=1}^C A_{ik}, \ D_{ij}=0\) if \(i\neq j\).\\
\midrule
\(\hat{A}\) & the normalized adjacency matrix of \(A\), \(\hat{A} = D^{-1}A\)\\
\midrule
\(\Delta\) & the graph Laplacian matrix, \(\Delta = D-A\)\\
\midrule
\(\tilde{\Delta}\) & the normalized Laplacian matrix, \(\tilde{\Delta} = I - D^{-1}A\)\\
\midrule
\(N\) & the number of nodes in a graph \(G\)\\
\midrule
\(M\) &  the size of the training set \\
\midrule
\(F\) & the dimensionality of node features\\
\midrule
\(C\) & the number of classes, as well as the dimensionality of model outputs\\
\midrule
\(H\) & the dimensionality of hidden units in GNN\\
\midrule
\(X\) & \(X \in \mathbb{R}^{N \times F}\), node features\\
\midrule
\(Z\) & \(Z \in \mathbb{R}^{N \times C}\), model outputs\\
\midrule
\(Z^\prime\) & the propagated model output, \(Z^\prime = \hat{A}Z\) \\
\midrule
\(P\) &  \(P \in \mathbb{R}^{N \times C}\), \textit{softmax} of \(Z\), predicted posterior class probability\\
\midrule
\(Q\) &  \(Q \in \mathbb{R}^{N \times C}\), \textit{softmax} of \(Z^\prime\)\\
\midrule
\(Y\) &  \(Y \in \mathbb{N}^{N \times C}\), one-hot ground truth label, \(Y_{ic}=1\) if  \(v_i\) is of class \(c\), otherwise \(Y_{ic}=0\)\\
\midrule
\(x_i\) & \(x_i \in \mathbb{R}^F\), the feature of \(v_i\), \(x_i = (X)_i^\top\)\\
\midrule
\(z_i\) & \(z_i \in \mathbb{R}^C\), the learned representation of \(v_i\), \(z_i = (Z)_i^\top\)\\
\midrule
\(y_i\) & \(y_i \in \mathbb{R}^C\), the \(i\)-th row vector of \(Y\)\\        \midrule
\(p_i\) & \(p_i \in \mathbb{R}^C\), the \(i\)-th row vector of \(P\)\\
\midrule
\((\cdot)^\top_i\) & the \(i\)-th row of \((\cdot)\)\\
\midrule
\(|S|\) & the size of a set \(S\)\\
\midrule
        \(\mathcal{L}_{cls}\) & the supervised classification loss, e.g., cross entropy loss\\
        \midrule
        \(\mathcal{L}_{P\text{-}reg}\) & the Propagation-regularization term  \\
        \bottomrule
    \end{tabular}
\end{table}

\section{Proofs to Lemmas and Theorems}\label{app:proofs}
\subsection{Proof of Theorem~\ref{th:sq-preg}}
\begin{proof} The squared-error P-reg  \( \mathcal{L}_{P\text{-}SE}\) is given as follows:
  \[
    \phi_{SE}(Z, \hat{A}Z) = \frac{1}{2}\sum_{i=1}^N\left\|(\hat{A}Z)_i^\top-\left(Z\right)_i^\top\right\|_2^2 = \frac{1}{2} \left\|\hat{A}Z-Z\right\|_F^2 = \frac{1}{2} \left\|\tilde{\Delta}Z\right\|_F^2 = \frac{1}{2}\left< Z, \tilde{\Delta}^\top\tilde{\Delta} Z \right>.
  \]
Thus, from the traditional Laplacian-based regularization perspective, the squared-error P-reg, \( \mathcal{L}_{P\text{-}SE}\), is equivalent to  taking  \(R = r(\tilde{\Delta}) = \tilde{\Delta}^\top\tilde{\Delta}\) as the regularization matrix in  the class of  Laplacian-based regularization functions, i.e., \(\left<Z, RZ\right>\) .
\end{proof}

\subsection{Proof of Lemma~\ref{lemma:idgcn}}
\begin{proof}
    \label{proof:idgcn}
    \citet{chung1997spectral} proved that for a connected graph with self-loops ($A:=A+I$), the largest eigenvalue \(\lambda_1\) of the normalized adjacency matrix \(\hat{A}=D^{-1}A\) is equal to \(2\), and the multiplicity of \(\lambda_1\) is 1. The corresponding eigenvector is \(u_1 = \mathbf{1} = \left(1,\ldots, 1\right)^\top\). And the smallest eigenvalue $\lambda_n$ is $0$. 
    
    Denote the \(j\)-th column vector of \(Z\) as \(Z_j, \ j\in [1, C]\). By applying the power iteration method, we obtain \(\mathbf{1} = u_1 =  \lim_{k\rightarrow \infty}\frac{\hat{A}^k Z_j}{\|\hat{A}^k Z_j\|} \), and thus \(\hat{A}^\infty Z_j = k \mathbf{1} \ \ \forall j \in [1, C], \text{ where } k \in \mathbb{R}\backslash\{0\}\) is a constant. Thus, we know the column vectors of \(\tilde{Z} = \hat{A}^\infty Z\) are all vector \(\mathbf{1}\) multiplied by constants, which means that all the row vectors of \(\tilde{Z}\) are the same. 
\end{proof}

\subsection{Proof of Lemma~\ref{lemma:preg-se}}
\begin{proof}\label{proof:preg-se}
    \citet{chung1997spectral} proved that for the Laplacian matrix \(\Delta\) of a connected graph \(G\), \(0\) is an eigenvalue of multiplicity 1 for \(\Delta\), and its corresponding eigenvector is \(\mathbf{1} = \left(1,\ldots, 1\right)^\top \in \mathbb{R}^N\). Then, if we take each column vector of \(\tilde{Z} \) as \(\tilde{Z}_j = k\mathbf{1}\), where \(j \in \left[1,\ldots, C\right]  \text{ and }  k \in \mathbb{R}\backslash\{0\}\) is a constant (then all row vectors of \(\tilde{Z}\) are the same), we have \(\Delta\tilde{Z} = \mathbf{0}\). Thus, \(\left\|\Delta\tilde{Z} \right\|_F^2 = \left\|D\tilde{Z} - A \tilde{Z} \right\|_F^2=0\), indicating  \(\phi_{SE}(Z, AZ) = \left\|\tilde{Z} - \hat{A}\tilde{Z} \right\|_F^2 = \left\|D^{-1}\left(D\tilde{Z} - A \tilde{Z}\right) \right\|_F^2= \sum_{i=1}^N\frac{1}{D_{ii}}\left\| \left(\tilde{\Delta}Z\right)_i^\top \right\|_2^2 =0\), which is the minimum value of \(\phi_{SE}\). Thus, such \(\tilde{Z}\) with all row vectors \(\tilde{z}_i \ \forall i \in [1, \ldots, N]\) being the same is an optimal solution to \(\arg \min_{Z} \left\|\hat{A}Z - Z \right\|_F^2\).
\end{proof}

\subsection{Proof of Theorem~\ref{theorem:preg-eq-idgcn}}
\label{proof:preg-eq-idgcn}
\begin{proof}
Define \(\tilde{Z} = \hat{A}^\infty Z\). We want to show that \(\tilde{Z}\) is the optimal solution for all the three versions of P-reg \(\phi(Z, \hat{A}Z)\).

Consider the Cross Entropy version first: \(\arg \min_Z \phi_{CE}(Z, \hat{A}Z) = -\sum_{i=1}^N \sum_{j=1}^{C}P_{ij}\log Q_{ij}\).

By computing the gradient of \(P\) to \(Z_{ij}\):
\[\sum_{m=1}^C\frac{\partial P_{im}}{\partial Z_{ij}}= \sum_{m=1}^C \left( -P_{ij}P_{im} \right) + P_{ij} =P_{ij}\left( 1- \sum_{m=1}^CP_{im} \right).\]
We can get \(\sum_{m=1}^C\frac{\partial P_{im}}{\partial Z_{ij}}=0\) because  \(\sum_{m=1}^C P_{im} = 1\).

Then we can compute the gradient of the Cross Entropy version of P-reg \(\frac{\partial \mathcal{L}}{\partial Z_{ij}} \) :
\[
    \begin{aligned}
    \frac{\partial \mathcal{L}}{\partial Z_{ij}} &= \sum_{k\in \mathcal{N}\left(i\right)\cup i} \frac{\partial \mathcal{L}_k}{\partial Z_{ij}} \\
    &= \frac{\partial \mathcal{L}_i}{\partial Z_{ij}} + \sum_{k\in \mathcal{N}\left(i\right)} \frac{\partial \mathcal{L}_k}{\partial Z_{ij}} \\
    &=-\frac{1}{N} \sum_{m=1}^C \log \left( \frac{1}{d_i} \sum_{k \in \mathcal{N}\left(i\right)} P_{km}\right) \frac{\partial P_{im}}{\partial Z_{ij}} + \sum_{k\in \mathcal{N}\left(i\right)} \frac{\partial \mathcal{L}_k}{\partial Z_{ij}} \\
    &= 0 + \sum_{k\in \mathcal{N}\left(i\right)} \frac{\partial \mathcal{L}_k}{\partial Z_{ij}}\\
    &= -\frac{1}{N} \sum_{k\in \mathcal{N}\left(i\right)} \sum_{m=1}^C \left( P_{km} \frac{1}{\sum_{l \in \mathcal{N}\left(k\right)\backslash i} P_{lm} + P_{im} } \right) \frac{\partial P_{im}}{\partial Z_{ij}} \\
    &=\! -\frac{1}{N} \sum_{k\in \mathcal{N}\left(i\right)} \! \left( \sum_{m=1}^C \! \left(  \frac{P_{km}}{\sum_{l \in \mathcal{N}\left(k\right)\backslash i} P_{lm}\! +\! P_{im} } \right) \left(-P_{ij}P_{im}\right)\! +\! \left(  \frac{P_{kj}}{\sum_{l \in \mathcal{N}\left(k\right)\backslash i}\! P_{lj}\! +\! P_{ij} } \right) P_{ij}  \right).
    \end{aligned}
\]

Consider the situation \(rank(Z)=1\). %
For \(P=\text{softmax}(Z)\), we know that all row vectors of \(P\) will be the same, i.e., \(P_{1m}= \cdots = P_{Nm} \ \ \forall m \in \left[1, C\right]\). Taking this into the above \(\frac{\partial \mathcal{L}}{\partial Z_{ij}}\), and considering \(\sum_{m=1}^{C}P_{im} = 1\),  we can get \(\frac{\partial \mathcal{L}}{\partial Z_{ij}} =-\frac{1}{N} \sum_{k\in \mathcal{N}\left(i\right)} \left( \sum_{m=1}^C \left(  \frac{1}{d_k} \right) \left(-P_{ij}P_{im}\right) + \left(  \frac{1}{d_k } \right) P_{ij}  \right) = -\frac{1}{N} \sum_{k\in \mathcal{N}\left(i\right)} \left(\left(  \frac{1}{d_k} \right) P_{ij} \sum_{m=1}^C \left(1-P_{im}\right)  \right)=  0\). Thus, the P-reg \(\phi_{CE}(Z, \hat{A}Z)\) is optimized when \(rank(Z)=1\). Taking \(\tilde{Z}=\hat{A}^\infty Z\), from Lemma~\ref{lemma:idgcn} we know \(rank(\tilde{Z})=1\). Thus, \(\tilde{Z}\) is the optimal solution for the  Cross Entropy version of P-reg.

And for the Squared Error version \(\arg \min_Z\phi_{SE}(Z, \hat{A}Z)=\frac{1}{2}\sum_{i=1}^N\left\|(\hat{A}Z)_i^\top-\left(Z\right)_i^\top\right\|_2^2\), taking \(\tilde{Z} =\hat{A}^\infty Z\),  we have \(\tilde{Z} = \hat{A}\tilde{Z}\). Thus,  \(\frac{1}{2}\sum_{i=1}^N\left\|(\hat{A}\tilde{Z})_i^\top-\left(\tilde{Z}\right)_i^\top\right\|_2^2=0\), which is the minimum value for \(\phi_{SE}(Z, \hat{A}Z)\). Thus, \(\tilde{Z}\) is the optimal solution to \(\phi_{SE}(Z, \hat{A}Z)\).

Similarly, consider the KL Divergence version \( \arg \min_Z\phi_{KL}(Z, \hat{A}Z) = \sum_{i=1}^N \sum_{j=1}^{C}P_{ij}\log \frac{P_{ij}}{Q_{ij}}\). When taking \(\tilde{Z} = \hat{A}^\infty Z\), we can get \(P=Q\) because  \(rank(\tilde{Z})=1\). Thus, \(\phi_{KL}(\tilde{Z}, \hat{A}\tilde{Z})=0\), which is the minimum value of \(\phi_{KL}(Z, \hat{A}Z)\). Thus \(\tilde{Z}\) is the optimal solution to \(\phi_{KL}(Z, \hat{A}Z)\).
\end{proof}

\section{Annealing and Thresholding of P-reg}
\label{app:anneal-thres}

\subsection{Annealing of P-reg}

In practice, we adopt a relative small \(\mu\) to make the classification loss dominate in the early stage. After \(f_1\) becomes a reasonably good predictor, the P-reg term will then dominate to further fine-tune the model parameters. A relatively small \(\mu\) will implicitly make P-reg work in the later stage of training.

To explicitly enforce the model to converge first and then let P-reg further tune the model parameters gradually, we could have an annealing version of P-reg as:
\[
  \mathcal{L} = \text{CrossEntropy}(Z, Y_{train}) +  \mu^t  \frac{1}{N}\phi\left(Z, \hat{A}Z\right),
\]
where \( 0< \mu < 1\), and \(t = \frac{1}{epoch}\), \(1 \leq epoch \leq E\), \(E\) is the total epochs.

\subsection{Thresholding of P-reg}

As it is shown in the analysis in Section~\ref{sec:preg-deepgcn} of the main paper, we do not expect the P-reg term to be exact $0$, i.e., $Z = \hat{A}Z$, since this causes over-smoothing and makes the representation for nodes in-discriminable.

Thus, we could strengthen P-reg by only computing its backward gradient when it is larger than a specified threshold. A hinge loss term can help achieve this:
\[
  \mathcal{L} = \text{CrossEntropy}\left(Z, Y_{train}\right) +  \mu \max \left(0,  \frac{1}{N}\phi\left(Z, \hat{A}Z\right) - \tau\right),
\]
where  \(\tau > 0\) is the threshold above which P-reg is applied.

\section{Related Work}
\label{app:related}

\paragraph{Prediction/Label Propagation.} In P-reg, \(\phi (Z, \hat{A}Z)\), where \(Z \in \mathbb{R}^{N\times C}\) is the output of a graph neural network \(f(A, X)\) and can be used for prediction,  \(\hat{A}Z\) can be regarded as to further propagate the prediction vectors. In classical Graph Label Propagation algorithms~\citep{zhu2003semi, zhou2003learning}, they directly propagate the labels of nodes in an iterative manner to predict over unlabeled nodes, i.e., \(Z_{t+1} = \alpha\hat{A}Z_t + \left(1-\alpha\right)Y\). And there are some adaptive variants~\citep{wang2007label, karasuyama2013manifold, liu2018learning} trying to build better graph structures from the node features for propagation. There are also some work~\citep{li2019label, wang2020unifying} attempting to unify graph neural networks with graph label propagation algorithms. \citet{li2019label} proposed Generalized Label Propagation, which uses the closed-form solution to Laplacian regularized label propagation as the graph filter for node features, and \citet{wang2020unifying} proposed to use the classification loss of label propagation as a regularizer to optimize the graph structure \(A\) to help graph convolution networks achieve higher node classification accuracy. In contrast, P-reg is a regularizer to minimize the gap between the trained graph neural network prediction outputs \(Z\) and the propagated prediction outputs \(\hat{A}Z\), which is different from the above-mentioned works. 

\paragraph{Soft Targets (Penalizing Label Confidence).}  Confidence Penalty~\citep{miller1996global} comes from the maximum entropy principle~\citep{jaynes1957information}. It adds the negative entropy of the network outputs to the classification loss as a regularizer, \(\mathcal{L}^{CP}=-\sum_{i \in \text{train}}y_i \log p_i - \beta \left (- \sum_{i} p_i \log p_i\right)\), where \(p_i\) is the output class probability distillation for the \(i\)-th sample, thus penalizing the confidence of the network prediction.  Label Smoothing~\citep{szegedy2016rethinking} is a strategy to improve the classification accuracy by softening the one-hot hard targets \(y_c=1,\ y_i=0 \ \forall i \neq c\) into \(y_i^{LS} = \left(1-\alpha\right)y_i + \nicefrac{\alpha}{C}\), where \(c\) is the correct label and \(C\) is the number of classes. Label Smoothing has been adopted in many state-of-the-art models~\citep{zoph2018learning, real2019regularized, huang2019gpipe, vaswani2017attention} in computer vision and natural language processing. \citet{pereyra2017regularizing} and \citet{rafael2019when} conducted extensive experiments on Label Smoothing and Confidence Penalty to show their effectiveness in improving the generalization of neural network models. Different from manually smoothing the targets, knowledge distillation~\citep{hinton2015distilling} trains a teacher model to generate soft targets for student models to learn. Using P-reg can be viewed as a self-distillation~\citep{zhang2019your} model, treating the trained 2-layer graph neural network model output as the soft supervision signal for all nodes' neighbors' aggregated output. And the whole model is trained in an end-to-end manner.

\paragraph{Graph Regularization.} Apart from the traditional graph regularization theory  based on Laplacian operators, which is to extend Laplacian regularizer by applying \(r\) to the spectrum of Laplacian \(\Delta\), i.e., \(r(\Delta)\)~\citep{civin1960linear}, some modern regularizers for graph neural networks have been proposed. GraphSGAN~\citep{ding2018semi} used generative adversarial networks to generate fake examples in low density region to help the semi-supervised learning model generalize better. GraphAT~\citep{feng2019graph}, as well as BVAT~\citep{deng2019batch}, incorporates the adversarial training methods into the training of GNN models, which adds adversarial perturbation into the input data, where adversarial perturbation is the perturbation along to the direction of the model gradient \(\frac{\partial f}{\partial x}\). GAM~\citep{otilia2019graph} co-trains GNN models with an additional agreement model which gives the probability that two nodes have the same label. GraphMix~\citep{verma2019graphmix} also uses the idea of co-training~\citep{blum1998combining}, by using a parameters-shared fully-connected network to make GNNs more generalizable. It also combined many semi-supervised techniques such as Mixup~\citep{zhang2017mixup}, entropy minimization with Sharpening~\citep{gradvalet2005semi}, Exponential Moving Average of predictions~\citep{tarvainen2017mean} and so on. These methods either need a pre-trained model to generated adversarial samples, or need additional gradient computation to modify the input data, or use too many semi-supervised learning tricks. In contrast, P-reg is a regularizer relying on no extra models or input modification, and P-reg has a low computation cost and memory overhead.

\section{More Details about the Experiments}
\label{app:exp-detail}
\subsection{Datasets}
Table~\ref{table:dataset_statistics} lists some statistics of the datasets used in our node classification experiments.
CORA, CiteSeer and PubMed are citation networks prepared by~\citet{yang2016revisiting}. \citet{shchur2019pitfalls} provided the pre-processed  Computers, Photo, CS and Photo datasets. The Computers and Photo datasets are Amazon co-purchase graphs from \citet{mcauley2015image}. The CS and Photo datasets are obtained from the Microsoft Academic Graph. All these datasets can be downloaded  from the PyTorch-geometric~\cite{pytorch-geometric}  built-in corresponding dataset loader.

\begin{table}[h]
	\caption{Node classification datasets}
	\label{table:dataset_statistics}
	\centering
	\begin{tabular}{l r r r r r }
		\toprule
		            &    Nodes &     Edges & Features  & Classes & \\
		\midrule
		CORA        &    2,708 &     5,278 &     1,433 &       7 &   \\
		CiteSeer    &    4,552 &     3,668 &     3,327 &       6 &    \\
		PubMed      &   19,717 &    44,324 &       500 &       3 &   \\
		CS          &   18,333 &    81,894 &     6,805 &      15 &   \\
		Physics     &   34,493 &   247,962 &     8,415 &       5 &    \\
		Computers   &   13,752 &   245,861 &       767 &      10 &    \\
		Photo       &    7,650 &  119,081  &       745 &       8 &    \\
		\bottomrule
	\end{tabular}
\end{table}

\subsection{Models}
In this section, we describe the details of the models we used in the node classification experiments.

The GCN has 2 graph convolutional layers, with the size of hidden units equal to 64.  \textit{ReLU} and a dropout layer with probability \(0.5\) are applied to the output of the first graph convolutional layer.

The GAT  has 2 graph attention layers. The first layer has 8 heads with the size of hidden units equal to 16. The second graph attention layer has 1 head. Both layers' attention dropout probabilities are \(0.6\). A dropout layer is applied before each graph attention layer, with dropout probability  \(0.6\). \textit{ELU} is used as the activation function.

The MLP has 2 fully connected layers, with the size of hidden units equal to 16. \textit{ReLU} is used as the activation function.

\subsection{Training and Evaluating}\label{app:training}
We used the Adam~\cite{kingma2014adam} optimizer with  learning rate equal to \(0.01\) to train all the models. For experiments on CORA, CiteSeer and PubMed, \(L_2\) weight decay of \(5\times 10^{-4}\) was applied, and for experiments on the CS, Physics, Computers and Photo datasets,  no weight decay was applied.  We used early stopping to determine the training epochs. If validation accuracy no longer increased for \(200\) epochs, the training was stopped. All these hyperparameters were fixed in all the experiments. 

The search space for  \(\mu\) is 20 values evenly chosen from \([0, 1]\) based on the validation accuracy. \(\mu\) remains the same for each cell (model \(\times\) dataset) in  Table~\ref{table:exp-results}.

To avoid bias and randomness, for the random split experiments, we evaluated each experiment on 5 randomly generated splits, and 5 different trials for each split. The training/validation/test splits for all datasets are 20/30/rest nodes per class. These settings follow the way of~\cite{shchur2019pitfalls}.

For the standard split experiments using the state-of-the-art methods, the code we used is either the official code released by the authors or from the PyTorch-geometric~\cite{pytorch-geometric}  implementation for benchmarking. Experimental results for all the models were from 10 random trials with initialization random seeds from 0 to 9.

\subsection{Software and Hardware}
P-reg in our experiments was implemented on PyTorch 1.5.0~\cite{pytorch} and PyTorch-geometric 1.4.3~\cite{pytorch-geometric}. We ran our experiments on Ubuntu 18.04 LTS with CUDA 10.1. The hardware platform we used in the experiments has 40 cores Intel(R) Xeon(R) Silver 4114 CPU @ 2.20GHz, 256 GB Memory, and 4 Nvidia RTX 2080Ti graphics cards.

\section{Choices of {$\phi$}}
The node classification accuracy of GCN and GAT using the three versions of P-reg, i.e., Squared Error, Cross Entropy, and KL Divergence on the 7 datasets using random splits is reported in Table~\ref{table:phi}. The three versions of P-reg have similar performance, and we report the results of the Cross Entropy version of P-reg in the main paper because it shows better stability.

\label{app:phi}
\begin{table}[h]
    \caption{The performance of P-reg with different $\phi$ functions}
    \label{table:phi}
    \centering
        \begin{tabular}{@{}cllllllll@{}}
        \toprule
        &                       & \multicolumn{1}{c}{CORA} & \multicolumn{1}{c}{CiteSeer} & \multicolumn{1}{c}{PubMed} & \multicolumn{1}{c}{CS} & \multicolumn{1}{c}{Physics} & \multicolumn{1}{c}{Computers} & \multicolumn{1}{c}{Photo} \\ \midrule
        \multirow{5}{*}{GCN} & Vanilla   & 79.47\PM{0.89} & 69.45\PM{0.99} & 76.26\PM{1.30} & 91.48\PM{0.42} & 93.66\PM{0.41} & 80.73\PM{1.52} & 89.24\PM{0.76} \\
        \cmidrule(l){2-9}
        & Squared Error   & 82.85\PM{1.08}  & 71.58\PM{1.86}  & 77.43\PM{1.63}  & 92.59\PM{0.33}  & 94.21\PM{0.69}  & 81.80\PM{1.53}  & 91.10\PM{0.73} \\
        & Cross Entropy & 82.83\PM{1.16}  & 71.63\PM{2.17}  & 77.37\PM{1.50}  & 92.58\PM{0.32}  & 94.41\PM{0.74}  & 81.66\PM{1.42}  & 91.24\PM{0.75} \\
        & KL Divergence  & 80.40\PM{0.99}  & 69.76\PM{1.15}  & 77.37\PM{0.68}  & 92.07\PM{0.50}  & 94.19\PM{0.39}  & 82.23\PM{1.34}  & 90.84\PM{1.02} \\
        \midrule
        \multirow{5}{*}{GAT} & Vanilla & 79.47\PM{1.77} & 69.28\PM{1.56} & 76.64\PM{2.03} & 91.15\PM{0.36} & 93.08\PM{0.55} & 81.00\PM{2.02} & 89.55\PM{1.20} \\
        \cmidrule(l){2-9}
        & Squared Error   & 82.96\PM{1.45}  & 70.60\PM{1.50}  & 76.62\PM{1.99}  & 91.83\PM{0.22}  & 94.14\PM{0.55}  & 76.89\PM{1.23}  & 86.53\PM{2.03} \\
        & Cross Entropy   & 82.97\PM{1.19}  & 70.00\PM{1.89}  & 76.56\PM{1.99}  & 91.92\PM{0.20}  & 94.28\PM{0.29}  & 83.68\PM{2.24}  & 91.31\PM{1.06} \\
        & KL Divergence    & 80.46\PM{1.22}  & 69.37\PM{1.76}  & 76.38\PM{1.85}  & 90.97\PM{0.33}  & 93.29\PM{0.42}  & 79.64\PM{2.77}  & 90.37\PM{1.23} \\
        \bottomrule
    \end{tabular}
\end{table}

\end{document}